\documentclass[]{fairmeta}

\usepackage{xspace}
\usepackage{amsmath,amsthm,amssymb,amsfonts}
\usepackage{algorithm}
\usepackage{algpseudocode}
\usepackage[endLComment=,italicComments=false]{algpseudocodex}
\usepackage{tikz-cd}
\usepackage[framemethod=TikZ]{mdframed}

\usepackage{amsmath,amsfonts,bm}

\def\eqref#1{equation~\ref{#1}}

\def\1{\bm{1}}

\DeclareMathAlphabet{\mathsfit}{\encodingdefault}{\sfdefault}{m}{sl}
\SetMathAlphabet{\mathsfit}{bold}{\encodingdefault}{\sfdefault}{bx}{n}

\def\gA{{\mathcal{A}}}

\def\gM{{\mathcal{M}}}

\def\gR{{\mathcal{R}}}
\def\gS{{\mathcal{S}}}

\def\gX{{\mathcal{X}}}

\newcommand{\E}{\mathbb{E}}

\DeclareMathOperator*{\argmin}{arg\,min}

\usepackage{macros}
\newcommand{\algname}{TD-JEPA\xspace}

\newmdenv[backgroundcolor=metabg, roundcorner=5pt, skipabove=7pt, linewidth=0pt, innertopmargin=4pt]{metaframe}

\newtheorem{proposition}{Proposition}

\newtheorem{theorem}{Theorem}

\newtheorem{remark}{Remark}

\definecolor{our_blue}{HTML}{0173B2}
\definecolor{our_bluel}{HTML}{80B9D8}
\definecolor{our_orange}{HTML}{DE8F05}
\definecolor{our_orangel}{HTML}{EEC782}
\definecolor{our_pink}{HTML}{CC78BC}

\usepackage{listings}
\lstdefinestyle{mypython}{
    language=Python,
    basicstyle=\ttfamily\scriptsize,
    keywordstyle=\color{our_blue}\bfseries,
    commentstyle=\color{gray}\itshape,
    numbers=left,
    numberstyle=\tiny\color{gray},
    backgroundcolor=\color{gray!10},
    tabsize=4,
    captionpos=b
}

\usepackage[titletoc,page]{appendix}
\usepackage{titletoc}
\usepackage{tocloft}

\title{TD-JEPA: Latent-predictive Representations for Zero-Shot Reinforcement Learning}

\author[1,2,3,*]{Marco Bagatella}
\author[1]{Matteo Pirotta}
\author[1]{Ahmed Touati}
\author[1]{Alessandro Lazaric}
\author[1]{Andrea Tirinzoni}

\affiliation[1]{FAIR at Meta}
\affiliation[2]{ETH Zurich}
\affiliation[3]{Max Planck Institute for Intelligent Systems, T\"{u}bingen}

\contribution[*]{Work done at Meta}

\abstract{Latent prediction--where agents learn by predicting their own latents--has emerged as a powerful paradigm for training general representations in machine learning.
In reinforcement learning (RL), this approach has been explored to define auxiliary losses for a variety of settings, including reward-based and unsupervised RL, behavior cloning, and world modeling. While existing methods are typically limited to single-task learning, one-step prediction, or on-policy trajectory data, we show that temporal difference (TD) learning enables learning representations predictive of long-term latent dynamics across multiple policies from offline, reward-free transitions.
Building on this, we introduce \algname, which leverages TD-based latent-predictive representations into unsupervised RL.
\algname trains explicit state and task encoders, a policy-conditioned multi-step predictor, and a set of parameterized policies directly in latent space. This enables zero-shot optimization of any reward function at test time.
Theoretically, we show that an idealized variant of \algname avoids collapse with proper initialization, and learns encoders that capture a low-rank factorization of long-term policy dynamics, while the predictor recovers their successor features in latent space. Empirically, \algname matches or outperforms state-of-the-art baselines on locomotion, navigation, and manipulation tasks across 13 datasets in ExoRL and OGBench, especially in the challenging setting of zero-shot RL from pixels.
}

\date{\today}
\correspondence{\email{tirinzoni@meta.com}}

\begin{document}

\maketitle

\section{Introduction}
\label{sec:intro}

Learning effective state representations is a core challenge in reinforcement learning (RL). Useful representations should capture the dynamics of the environment in a way that supports efficient value estimation and policy optimization across tasks \citep{watter2015embed, silver2018general, hafner2019learning, gelada2019deepmdp}. A promising line of work is latent-predictive (a.k.a. self-predictive) representation learning \citep{schwarzer2020data, grill2020bootstrap,guo2020bootstrap,tang2023understanding}, an instance of the joint-embedding predictive architecture~\citep[][JEPA]{lecun2022path} paradigm.
These algorithms jointly learn a \emph{state encoder} $\phi(s)$ and a \emph{predictor} $P$, i.e., a latent dynamics model estimating the representation of a future state $s'$: $P(\phi(s)) \simeq \phi(s')$. Latent-predictive methods thus perform \emph{self-supervised} learning entirely in latent space without any reward or reconstruction of (possibly high-dimensional) states. 

Several RL methods leverage latent prediction as an auxiliary loss to improve sample efficiency and generalization in reward-based learning~\citep{schwarzer2020data,guo2020bootstrap,hansen2023td}, behavior cloning~\citep{lawson2025self}, and curiosity-driven exploration~\citep{guo2022byol}. As latent-predictive losses do not require any reward, they have been recently used for unsupervised RL: \citet{assran2025vjepa2selfsupervisedvideo}, \citet{zhou2025dinowmworldmodelspretrained} and \citet{sobal2025learning} learn latent world models that can solve goal-reaching tasks via test-time planning, whereas \citet{jajoo2025regularized} learn a state encoder from trajectory data to define the space of tasks used to optimize zero-shot unsupervised policies.

\begin{figure}
    \centering
    \vspace{-3mm}
    \includegraphics[width=\linewidth]{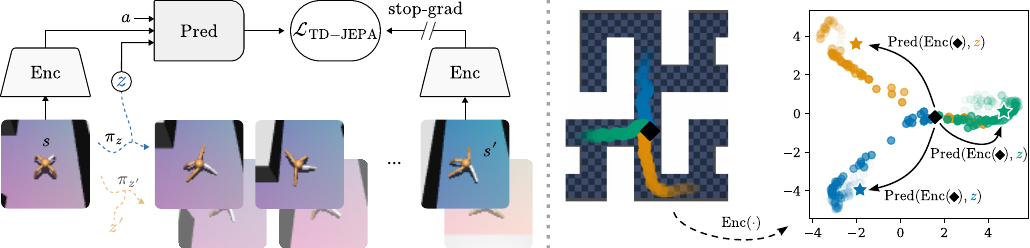}
    \vspace{-5mm}
    \caption{\algname trains policies $\pi_z$ parameterized by latents $z$. The predictor, conditioned on $z$, predicts the representations of future states visited by $\pi_z$ (\emph{left}). When trained via TD, the predictor (arrows on the \textit{right}) approximates successor features for each policy, i.e., the weighted barycenter (stars) of representations of visited states (circles).
    }
    \vspace{-5mm}
    \label{fig:teaser}
\end{figure}

This paper proposes a novel way to instantiate latent-predictive representations for unsupervised RL. While previous methods have largely focused on either one-step dynamics, single-task/single-policy training, or relied on on-policy data, we introduce a policy-conditioned, multi-step formulation based on a novel off-policy temporal-difference loss. This objective encourages representations that are predictive not only of immediate transitions, but also of long-term features relevant for value estimation across multiple policies. This property makes such representations and the associated predictors particularly well-suited for integration with off-policy, successor-feature based approaches to zero-shot unsupervised RL \citep{touati2021learning, touati2023does, park2024foundation}. 

We thus instantiate temporal difference latent-predictive representation learning into \algname, a
zero-shot unsupervised RL algorithm which pre-trains four components: a state encoder, a policy-conditioned multi-step predictor, a task encoder, and a set of parameterized policies, all of which are learned end-to-end from offline, reward-free transitions.
Departing from previous approaches, latent prediction is not merely an auxiliary loss, but rather the core objective that enables \algname to learn all the components needed to distill zero-shot policies.
In fact, the predictor may be leveraged as an approximation of successor features (see Figure \ref{fig:teaser}) to extract policies mapping encoded observations to optimal actions for all reward functions in the span of the learned features.
This enables \algname to perform zero-shot policy optimization for any downstream reward, \textit{entirely in latent space}.

Theoretically, for an idealized version of \algname with linear predictors, we show that \textbf{1)} the representations do not collapse with a suitable initialization; \textbf{2)} they recover a low-rank factorization of the successor measures of the trained policies, while the predictor approximates successor features in latent space; \textbf{3)} they minimize an upper bound on the policy evaluation error for any reward, thus making zero-shot optimization possible. These results build on a novel ``gradient matching'' argument that extends and generalizes existing theoretical analyses of latent-predictive representations, and connect \algname with other unsupervised RL methods such as forward-backward~\citep[][]{touati2021learning} and intention-conditioned value functions~\citep[][]{ghosh2023reinforcement}.

Empirically, we evaluate \algname on 65 tasks across 13 datasets from ExoRL~\citep{Yarats2022exorl} and OGBench~\citep{Park25ogbench}, covering locomotion, navigation, and manipulation with both proprioceptive and pixel-based observations. \algname matches or outperforms state-of-the-art zero-shot baselines across these settings, in particular when learning from pixels, which has proven to be one of the most challenging settings for unsupervised RL so far. Moreover, we ablate several dimensions of the algorithm, demonstrating the importance of learning representations that are predictive of multi-step policy-dependent dynamics, and the advantage of training distinct state and task encoders. Finally, we show that learned representations can be easily reused
for offline or online RL, improving over zero-shot policies and learning from scratch.

\section{Preliminaries}\label{sec:prelims}

We consider a reward-free Markov Decision Process $\gM = (\gS, \gA, P, \gamma)$, where $\gS$ and $\gA$ are state and action spaces, $P$ is the probability measure over next states when taking action $a$ in state $s$ as $P(\mathrm{d}s' \mid s,a)$, and $\gamma \in [0, 1)$ is a discount factor.
Executing a Markov policy $\pi: \gS \to \mathrm{Prob}(\gA)$ induces an unnormalized distribution over visited states, which is referred to as the \textit{successor measure}:
\begin{equation}
    M^\pi(\gX \mid s, a) = \sum_{t=0}^\infty \gamma^t \mathrm{Pr}(s_{t+1} \in \gX | s, a, \pi) \quad \forall \; \gX \subseteq \gS.
\end{equation}
Given a reward function $r: \gS \to \mathbb{R}$ and a policy $\pi$, the action-value function $Q_r^\pi(s,a)$ measures the cumulative discounted reward obtained by the policy over an infinite horizon, i.e., $Q_r^\pi(s,a) = \E \big[ \sum_{t=0}^\infty \gamma^t r(s_{t+1})  \mid s, a, \pi \big]$. Action-value functions are connected to successor measures via
\begin{equation}\label{eq:q.and.m}
    Q_r^\pi(s,a) = \int_{s^+ \in \gS} M^\pi(\mathrm{d}s^+ \mid s, a) r(s') = \E_{s^+\sim M^\pi(\cdot|s,a)} \big[ r(s^+)\big], 
\end{equation}
which shows a convenient linear decomposition of $Q_r^\pi$ into the reward function and the dynamics induced by $\pi$. Standard RL agents aim at finding reward-maximizing policies $\pi_r^\star(s) \in \arg\max_{a\in\gA} Q^\star_r(s,a)$, 
where $Q^\star_r(s,a) := \max_\pi Q^\pi_r(s,a)$.

\textbf{Latent-predictive representations.} 
In high-dimensional settings, \emph{state encoders} $\phi: \gS \rightarrow \mathbb{R}^{d_\phi}$ may be learned to ease the estimation of action-value functions.
For instance, if an encoder $\phi$ is such that $Q_r^\pi(s,a) = \phi(s)^\top w_{a,r}^\pi$ for some vector $w_{a,r}^\pi\in\mathbb{R}^{d_\phi}$, then the RL process reduces to learning vectors in $\mathbb{R}^{d_\phi}$ rather than high-dimensional functions $Q_r^\pi(s,a)$. Latent-predictive learning has been shown to be an effective approach for this problem. In the simplest formulation, latent-predictive representations capture the one-step latent dynamics of a policy $\pi$ by minimizing the loss
\begin{equation}\label{eq:1step.sp.mc}
 \mathcal{L}_{\text{one-step}}(\phi, T) = \E_{s\sim\rho, a\sim \pi(\cdot|s), s'\sim P(\cdot|s,a)} \big[ \| T(\phi(s)) - \sg{\phi(s')}\|^2 \big],
\end{equation}
where $T: \mathbb{R}^{d_\phi} \rightarrow \mathbb{R}^{d_\phi}$ is a (possibly non-linear) predictor of the latent one-step dynamics induced by $\phi$ and policy $\pi$, and $\sg{\phi}$ denotes stop-gradient. Notably, optimizing for this loss does not require any decoding or reconstruction, and it only relies on an \emph{unsupervised} dataset $\mathcal{D}=\{(s,a,s')\}$. Different instantiations of this approach have been shown both empirically and theoretically to produce representations that accurately approximate action-value functions or policies \citep{guo2022byol,tang2023understanding,voelcker2024does,lawson2025self, fujimoto2025towards}.

\textbf{Successor-features and zero-shot unsupervised RL.} Considering a state encoder $\psi: \gS\rightarrow \mathbb{R}^{d_\psi}$ and the associated space of linear rewards $\gR_{\psi} = \{r(s) = \psi(s)^\top z \mid z \in \mathbb{R}^{d_\psi}\}$, Q-values for any reward function $r(s) = \psi(s)^\top z_r \in  \mathcal{R}_{\psi}$ can be written as
\begin{equation}\label{eq:q-vs-sf}
    Q_{r}^\pi(s,a) = \int_{s^+ \in \gS} M^\pi(\mathrm{d}s^+ \mid s, a) \psi(s^+)^\top z_r =\mathbb{E}_{s^+\sim M^\pi(\cdot|s,a)}\big[\psi(s^+)\big]^\top z_r :=  F_\psi^\pi(s,a)^\top z_r, 
\end{equation}
where $F_\psi^\pi(s,a)\in\mathbb{R}^{d_\psi}$ captures the \emph{successor features} of $\pi$ \citep{barreto2017successor}. The majority of unsupervised zero-shot RL methods~\citep{touati2021learning,park2024foundation,agarwal2024proto,jajoo2025regularized} learn successor features $F(s,a;z) \approx F^{\pi_z}_{\psi}(s,a)$ for a set of parameterized policies $\{\pi_z(s)\}_{z\in\mathcal{Z}}$, with $\mathcal{Z}\subseteq\mathbb{R}^d$, that are trained to be optimal for all rewards in $\mathcal{R}_\psi$, i.e., $\pi_z(s)\approx \arg\max_a F(s,a;z)^\top z$, where $F(s,a;z)^\top z$ is an approximation of $Q^\star_r(s,a)$ for $r(s)=\psi(s)^\top z$. 
At test time, given a reward function $r$, a vector $z_r\in\mathbb{R}^{d_\psi}$ is first obtained by projecting $r$ onto $\gR_\psi$, and the associated policy $\pi_{z_r}$ is then returned. 

Given the role played by $\psi$ in defining the space of tasks of interest, with an abuse of terminology, we will refer to $\psi$ as a \emph{task encoder}.
On the other hand, we shall call \emph{state encoder} a map $\phi: \gS \rightarrow \mathbb{R}^{d_\phi}$ that is used to embed states before feeding them into different networks (e.g., we will train successor features $F_\psi^\pi(\phi(s),a)$ and policies $\pi(\phi(s))$ in the latent space given by $\phi$).
While the zero-shot methods cited so far train the task encoder $\psi$ in different ways, and do not train any explicit state encoder $\phi$, the next section will show how multi-step policy-dependent latent-predictive learning can be used to train both simultaneously.

\section{Latent-Predictive Temporal-Difference Representations}

We begin by showing how the latent-predictive loss of Eq. \ref{eq:1step.sp.mc} can model multi-step and policy-dependent dynamics, and how temporal difference (TD) learning allows learning from offline transition data. We will then expand this idea to learn separate state and task embeddings, and finally show how it can be instantiated as a zero-shot unsupervised RL method. 

\subsection{Multi-step policy-conditioned latent prediction}\label{sec:sptd-symm}

Let $\{\pi_z\}_{z\in\mathcal{Z}}$ be a family of policies parameterized by $z\in\mathcal{Z}$, and $\mathcal{D} = \{(s,a,s')\}$ be a dataset of transitions. We train a state encoder $\phi: \gS \to \mathbb{R}^{d_\phi}$ and a \emph{policy-dependent} predictor $T_\phi: \mathbb{R}^{d_\phi} \times \gA \times \mathcal{Z}\to \mathbb{R}^{d_\phi}$ to be latent-predictive of the \emph{long-term} dynamics of the policies $\{\pi_z\}$, i.e., 
\begin{equation}\label{eq:mc-obj}
 \mathcal{L}_{\text{MC-JEPA}}(\phi, T_\phi) = \E_{(s,a)\sim\mathcal{D}, z\sim \mathcal{Z}, s^+\sim M^{\pi_z}(\cdot|s,a)} \big[ \| T_\phi(\phi(s),a,z) - \sg{\phi(s^+)}\|^2 \big],
\end{equation}
where \textit{MC-JEPA} stands for Monte-Carlo (MC) JEPA loss, as on-policy samples $s^+\sim M^{\pi_z}(\cdot|s,a)$ are needed for all policies of interest. Intuitively, $T_\phi(\phi(s),a,z)$ tries to predict future latent states visited by the policy $\pi_z$. More formally, predictors trained via minimization of $\mathcal{L}_{\text{MC-JEPA}}(\phi, T_\phi)$ approximate the successor features of $\phi$ in the latent space induced by $\phi$ itself.
\begin{metaframe}
\begin{proposition}\label{prop:pred.succ.features}
For any $\phi$ and $T_\phi$, we have the following equivalence 
\begin{equation}
    \mathcal{L}_{\text{MC-JEPA}}({\phi}, T_\phi) = \E_{(s,a)\sim\mathcal{D}, z\sim \mathcal{Z} } \big[ \| T_\phi({\phi}(s),a,z) - \sg{F_\phi^{\pi_z}(s,a)}\|^2 \big] + \mathrm{const}.
\end{equation}
\end{proposition}
\end{metaframe}
Given the connection between Q-functions and successor features (Eq. \ref{eq:q-vs-sf}), this result crucially relates multi-step latent prediction with value estimation across multiple policies. More precisely, it implies that the predictor enables policy evaluation and optimization of rewards in the span of $\phi$, as we detail at the end of this section. Since $F_\phi^{\pi_z}$ is the successor features of $\phi$, with the terminology introduced in Sec.~\ref{sec:prelims}, $\phi$ is used both as a \emph{state encoder}, i.e., to embed states passed to the predictor, and as a \emph{task encoder}, i.e., defining a space of reward functions.

Unfortunately, this loss cannot be estimated on off-policy data since it requires sampling from the successor measures of the given policies.
We can however leverage the previous result and the fact that successor features admit a Bellman equation $F_\phi^{\pi_z}(s,a) = \bE_{s'\sim P(\cdot | s,a), a'\sim\pi_z(s')}[\phi(s') + \gamma F_\phi^{\pi_z}(s',a')]$ \citep{barreto2017successor} to define a temporal-difference version of the previous loss:
\begin{equation}\label{eq:td-obj}
 \mathcal{L}_{\text{\algname}}(\phi, T_\phi) = \E_{(s,a,s')\sim\mathcal{D}, z\sim \mathcal{Z}, a'\sim \pi_z(\cdot|s')} \big[ \| T_\phi(\phi(s),a,z) - \sg{\phi(s')} - \gamma \sg{T_\phi(\phi(s'),a',z)}\|^2 \big].
\end{equation}
Unlike the Monte Carlo loss of Eq. \ref{eq:mc-obj}, $\mathcal{L}_{\text{\algname}}$ only requires sampling one-step transitions and actions from the given policies, and it can thus be estimated from off-policy, offline datasets.
\subsection{Training separate state and task representations}

\textcolor{black}{While in Eq. \ref{eq:mc-obj} and \ref{eq:td-obj} the same encoder $\phi$ is used for both state and task representations, these need not be the same in practice. Consider, for instance, a robot navigating a building: useful state representations may capture low-level dynamical information critical for control (e.g., joint positions and velocities), while task representations could abstract higher-level contextual features, such as the building's topology. In this case, a single representation might be either too complex, or too abstract: having flexibility over the dimensionality and content of each representation would be desirable.}
We thus now introduce an asymmetric variant that trains a distinct encoder $\psi: \gS \rightarrow \mathbb{R}^{d_\psi}$ to define the set of reward functions of interest (i.e., as a \textit{task} encoder). We first redefine the predictor as $T_\phi: \mathbb{R}^{d_\phi} \times \gA \times \mathcal{Z}\to \mathbb{R}^{d_\psi}$ and the latent-predictive Monte-Carlo loss to train $\phi$ and $T_\phi$ as
\begin{equation}\label{eq:mc-fw-sp-loss}
 \mathcal{L}_{\text{MC-JEPA}}(\phi, T_\phi, \psi) = \E_{(s,a)\sim\mathcal{D}, z\sim \mathcal{Z}, s^+\sim M^{\pi_z}(\cdot|s,a)} \big[ \| T_\phi(\phi(s),a,z) - \sg{\psi(s^+)}\|^2 \big],
\end{equation}
such that $T_\phi$ maps states encoded through $\phi$ to the long-term dynamics of a policy $\pi_z$ in the latent space induced, this time, by $\psi$. Similar to Prop.~\ref{prop:pred.succ.features}, $T_\phi$ approximates the successor features $F_\psi^{\pi_z}(s,a)$ of $\psi$ in the latent space induced by $\phi$. Symmetrically, we train $\psi$ together with an additional predictor $T_\psi: \mathbb{R}^{d_\psi} \times \gA \times \mathcal{Z}\to \mathbb{R}^{d_\phi}$. To do so, we follow existing literature -- according to which joint representations should be predictive of each other \citep{guo2020bootstrap,tang2023understanding} -- and train $\psi$ and $T_\psi$ through the same latent-predictive loss with the roles of $\phi$ and $\psi$ inverted, i.e., $\mathcal{L}_{\text{MC-JEPA}}(\psi, T_\psi, \phi)$.\footnote{While some existing works use forward-in-time sampling to train one representation and backward-in-time for the other, we use two forward-in-time losses. We further discuss this difference in App.~\ref{app:theory-general}.} As before, we can then design an off-policy TD variant of this loss,
\begin{equation}\label{eq:asymm.sptd.phi}
 \mathcal{L}_{\text{\algname}}(\phi, T_\phi, \psi) = \E_{\substack{(s,a,s')\sim\mathcal{D} \\z\sim \mathcal{Z}, a'\sim \pi_z(\cdot|s')}} \big[ \| T_\phi(\phi(s),a,z) - \sg{\psi(s')} - \gamma \sg{T_\phi(\phi(s'),a',z)}\|^2 \big],
\end{equation}
so that $\phi$ and $T_\phi$ are optimized via $\mathcal{L}_{\text{\algname}}(\phi, T_\phi, \psi)$, while $\psi$ and $T_\psi$ via $\mathcal{L}_{\text{\algname}}(\psi, T_\psi, \phi)$. %

\subsection{\algname representations for zero-shot RL}\label{sec:sptd-zero-shot}

\begin{algorithm}[t!]
    \caption{\algname for zero-shot RL}\label{alg:sptd}
    \begin{algorithmic}
    \footnotesize
    \State{\textbf{Inputs}: Dataset $\mathcal{D}$, batch size $B$, regularization coefficient $\lambda$, networks $\pi$, $T_\phi$, $\phi$, {$T_\psi$, $\psi$}}
    \State Initialize target networks: $T_\phi^{-}  \leftarrow T_\phi$, $\phi^- \leftarrow \phi$, { $T_\psi^{-}  \leftarrow T_\psi$, $\psi^- \leftarrow \psi$}
    \While{not converged}
    \State \(\triangleright\) \textcolor{gray}{Sample training batch}
    \State $\{(s_i, a_i, s_{i}')\}_{i=1}^B \sim \mathcal{D}$, $\{z_i \}_{i = 1}^B \sim \mathcal{Z}$, $\{a'_i\}_{i=1}^B \sim \{\sg{\pi(\phi^-(s_i'),z_i)}\}_{i=1}^B$
    \vspace{5pt}
    \State \vspace{0.3pt} \(\triangleright\) \textcolor{gray}{Compute latent-predictive losses}
    \State $\wh{\cL}_{\text{\algname}} (\phi, T_\phi, \psi) =\frac{1}{2 B} \sum_{i} \left\| T_\phi(\phi(s_i), a_i, z_i) - \sg{\psi^-(s_i')}  - \gamma \sg{T_\phi^-(\phi^-(s_i'),a_i',z_i)}  \right\|^2$
    \State {$\wh{\cL}_{\text{\algname}} (\psi, T_\psi, \phi) =\frac{1}{2 B} \sum_{i} \left\| T_\psi(\psi(s_i), a_i, z_i) - \sg{\phi^-(s_i')}  - \gamma \sg{T_\psi^{-}(\psi^-(s_i'),a_i',z_i)}  \right\|^2$}
    \State \vspace{0.3pt} \(\triangleright\) \textcolor{gray}{Compute orthonormality regularization losses}
    \State $\wh{\cL}_{\rm{REG}}(\phi) = \frac{1}{2 B (B-1)} \sum_{i \neq j} (\phi(s_i)^\top \phi(s_j))^2 - \frac{1}{B} \sum_{i}\phi(s_i)^\top \phi(s_i) $
    \State {$\wh{\cL}_{\rm{REG}}(\psi) = \frac{1}{2 B (B-1)} \sum_{i \neq j} (\psi(s_i)^\top \psi(s_j))^2 - \frac{1}{B} \sum_{i}\psi(s_i)^\top \psi(s_i)$}
    \State \vspace{0.3pt} \(\triangleright\) \textcolor{gray}{Compute actor loss}
    \State $\{\hat{a}_i\}_{i=1}^{B} \sim \{\pi(\phi(s_i),z_i)\}_{i=1}^B$
    \State $\wh{\cL}_{\rm{actor}} (\pi) = -\frac{1}{B}\sum_{i=1}^B T_\phi(\phi(s_i), \hat a_i, z_i)^\transp z_i $
    \State \vspace{0.1pt} Update $\phi$, $T_\phi$ to minimize $\wh{\cL}_{\text{\algname}} (\phi, T_\phi, \psi) + \lambda \wh{\cL}_{\rm{REG}}(\phi)$
    \State {Update $\psi$, $T_\psi$ to minimize  $\wh{\cL}_{\text{\algname}} (\psi, T_\psi, \phi) + \lambda \wh{\cL}_{\rm{REG}}(\psi)$}
    \State Update $\pi$ to minimize $\wh{\cL}_{\rm{actor}} (\pi)$
    \State Update target networks $\phi^-$, $T_\phi^{-}$, {$\psi^-$, $T_\psi^{-}$} via EMA of $\phi$, $T_\phi$, {$\psi$, $T_\psi$}
    \EndWhile
    \end{algorithmic}
\end{algorithm}

The relationship between the learned predictors and successor features suggests a seamless instantiation of \algname as a zero-shot unsupervised RL algorithm. Redefining the policy parameter space $\cZ$ as the task embedding space (i.e., $\mathcal{Z} \subseteq \mathbb{R}^{d_\psi}$), we train latent policies such that $\pi_z(\phi(s)) = \mathop{\text{argmax}}_a T_\phi(\phi(s),z,a)^\top z$ for all $z\in\mathcal{Z}$\footnote{\textcolor{black}{This decision additionally grounds $\psi$ as task encoder, and breaks the symmetry that could arise from the two encoders $\phi$ and $\psi$ being trained through similar latent-predictive objectives.}}. Since $T_\phi(\phi(s),z,a) \simeq F_\psi^{\pi_z}(s,a)$ (Proposition \ref{prop:pred.succ.features}), this produces optimal policies for all rewards in the span of $\psi$, learned directly from state representations $\phi(\cdot)$.
At test time, given an inference dataset of rewarded samples $\mathcal{D}_{\text{rwd}}=\{(s,r)\}$, the optimal policy $\pi_{z_r}$ can be retrieved by computing $z_r$ through linear regression, e.g. through the closed-form solution $z_r = \text{argmin}_z \, \mathbb{E}_{(s, r) \sim \mathcal{D}_{\text{rwd}}}[(r - \psi(s)^\top z)^2] = \bE_{s \sim \mathcal{D}_{\text{rwd}}}[\psi(s)\psi(s)^\transp]^{-1}\bE_{(s, r) \sim \mathcal{D}_{\text{rwd}}}[\psi(s)r(s)]$.
Alg.~\ref{alg:sptd} describes \algname, which combines $\mathcal{L}_{\text{\algname}}$ with 
stabilization strategies, e.g. target networks and covariance regularization.
We remark that latent prediction is not auxiliary: it is the core objective that trains encoders and predictors, from which zero-shot policies can be directly distilled.

\section{Theoretical Analysis}
\label{sec:theory}

We now provide some theoretical arguments showing how latent-predictive temporal difference representations capture the long-term dynamics of a given set of policies in a way that makes them amenable to zero-shot RL. Following \citet{tang2023understanding}, we consider a simplified tabular setting with linear predictors. We view the representation $\phi$ (resp. $\psi$) as a $S \times d_\phi$ (resp. $S \times d_\psi$) matrix, and consider action-free predictors $T_{\phi,z}$ (resp. $T_{\psi,z}$) as $d_\phi \times d_\psi$ (resp. $d_\psi \times d_\phi$) matrices for all $z$. The expression $T_\phi(\phi(s), a, z)$ in Eq.~\ref{eq:mc-fw-sp-loss} and \ref{eq:asymm.sptd.phi} thus reduces to $T_{\phi,z}^\transp \phi(s)$, while $M^\pi(s' | s,a)$ and $P(s' | s,a)$ are replaced by $M^{\pi_z}(s'|s) = M^{\pi_z}(s' | s, \pi_z(s))$ and $P^{\pi_z}(s' | s) = P(s'|s,\pi_z(s))$.

\subsection{Monte-Carlo losses}
We define a (non-latent-predictive) successor measure approximation loss
\begin{align}\label{eq:lr-reconstruction}
    \cL_{\mathrm{SM}}(\phi, \{T_z\}_z, \psi) :&= \frac{1}{2}\bE_{z \sim \cZ} \| \phi T_z \psi^\transp - M^{\pi_z} \|_F^2.
\end{align}
Minimizing $\cL_{\mathrm{SM}}$ is equivalent to finding the best multilinear approximation to the successor measures $M^{\pi_z}$. We prove the following connection with the Monte Carlo latent-predictive loss of Eq.~\ref{eq:mc-fw-sp-loss}.

\begin{metaframe}
\begin{theorem}\label{th:mc-sp-vs-recon}
    For fixed $\phi$ and $\psi$, let $T_{z}^\star, T_{\phi,z}^\star, T_{\psi,z}^\star$ be the optimal predictors for $\cL_{\mathrm{SM}}(\phi, T_z, \psi)$ (Eq.~\ref{eq:lr-reconstruction}), $\cL_{\text{MC-JEPA}}(\phi, T_{\phi,z}, \psi)$, $\cL_{\text{MC-JEPA}}(\psi, T_{\psi,z}, \phi)$ (Eq.~\ref{eq:mc-fw-sp-loss}), respectively. If (A1) $\phi^\transp \phi = \psi^\transp \psi = I$, (A2) the state distribution is uniform, and (A3) for all $z\in\cZ$, the matrix $P^{\pi_z}$ is symmetric, then
    \begin{enumerate}
        \item for all $z$, $\phi T_z^\star = \phi T_{\phi,z}^\star = \Pi_\phi M^{\pi_z} \psi$ and $\psi T_{\psi,z}^\star =  \psi (T_{z}^\star)^\transp =  \Pi_\psi M^{\pi_z} \phi$, where $\Pi_\phi$ (resp. $\Pi_\psi$) is an orthogonal projection on the span of $\phi$ (resp. $\psi$);
        \item $\nabla_\phi \cL_{\text{MC-JEPA}}(\phi, T_z, \psi) = \nabla_\phi \cL_{\mathrm{SM}}(\phi, T_z, \psi)$ and $\nabla_\psi \cL_{\text{MC-JEPA}}(\psi, T_z, \phi) = \nabla_\psi \cL_{\mathrm{SM}}(\phi, T_z^\transp, \psi)$.
    \end{enumerate}
\end{theorem}
\end{metaframe}

This result reveals that \textbf{1)} the optimal predictors for the successor measure loss $\cL_{\mathrm{SM}}$ and the latent-predictive loss $\cL_{\text{MC-JEPA}}$ match, and yield an orthogonal projection of the successor features $M^{\pi_z}\psi$ onto the $\phi$ space; \textbf{2)} the gradients w.r.t. the representations $\phi$ and $\psi$, when evaluated at any predictor, match among these two losses, showing that gradient descent on $\mathcal{L}_{\text{MC-JEPA}}$ would update representations in the direction that reduces $\cL_{\mathrm{SM}}$, hence improving the approximation of the successor measures.
This result follows as a special case of a novel theorem (see App.~\ref{app:theory-general}) generalizing and implying all previous guarantees for latent-predictive representations~\citep{tang2023understanding,khetarpal2024unifying,voelcker2024does,lawson2025self}, which we believe is of independent interest. Finally, we remark that, while the assumptions A1-A3 have been considered in all these related works, they can be relaxed, at the price of more involved proofs and notation, as shown in App.~\ref{app:theory-general}.

\subsection{Temporal-difference losses} We first derive a non-collapse guarantee. While a similar result was originally proved by~\citet{tang2023understanding} for the one-step loss (Eq.~\ref{eq:1step.sp.mc}), our case is more complex since TD latent-prediction can be seen as ``doubly latent-predictive'' (cf. Eq.~\ref{eq:asymm.sptd.phi}): $T_{\phi,z}^\transp \phi(s)$ is optimized to match a representation being learned -- $\psi(s^+)$ -- plus a bootstrapped version of itself -- $T_{\phi,z}^\transp \phi(s^+)$.
\begin{metaframe}
\begin{theorem}\label{th:td-constant-cov}
    Let $\phi_t$ and $\psi_t$ be the representations learned under a continuous-time relaxation of Eq.~\ref{eq:asymm.sptd.phi} where, at each step $t$, the optimal predictors for ($\phi_t$, $\psi_t$) are first computed and then a gradient step on ($\phi_t$, $\psi_t$) is taken (see App.~\ref{app:proof-non-collapse} for the explicit formulation). Then, the covariance matrices $\phi_t^\transp \phi_t$ and $\psi_t^\transp \psi_t$ are constant over time, i.e., $\phi_t^\transp \phi_t = \phi_0^\transp \phi_0$ and $\psi_t^\transp \psi_t = \psi_0^\transp \psi_0$ for all $t \geq 0$.
\end{theorem}
\end{metaframe}
This result suggests that, if predictors are trained at a faster rate than representations, the overall dynamics preserve their covariance, thus preventing $\phi$ and $\psi$ from collapsing to trivial solutions (e.g., $\phi=\psi=0$) when properly initialized, e.g., with unitary covariance.

As done for MC objectives (Th.~\ref{th:mc-sp-vs-recon}), we now show that the latent-predictive loss of \algname is related to forward and backward TD losses for approximating the successor measure \citep{blier2021learning}. \looseness -1
\begin{metaframe}
\begin{theorem}\label{th:td-sp-vs-den-new}
Consider the following TD losses for approximating the successor measure
    \begin{align}
        \cL_{\mathrm{fw}}(\phi, T_z, \psi) &:= \frac{1}{2}\bE_{z \sim \cZ}\left[ \| \phi T_z \psi^\transp - P^{\pi_z} - \gamma \sg{P^{\pi_z} \phi T_z \psi^\transp} \|_F^2 \right], \label{eq:td-den-fw}
        \\ \cL_{\mathrm{bw}}(\phi, T_z, \psi) &:= \frac{1}{2}\bE_{z \sim \cZ}\left[ \| \psi T_z \phi^\transp - (P^{\pi_z})^\transp - \gamma (P^{\pi_z})^\transp \sg{\psi T_z \phi^\transp} \|_F^2 \right]. \label{eq:td-den-bw}
    \end{align}
    For fixed $(\phi, \psi)$, let $T_{z, \mathrm{fw}}^\star, T_{z, \mathrm{bw}}^\star, T_{\phi,z}^\star, T_{\psi,z}^\star$ respectively be the optimal predictors for $\cL_{\mathrm{fw}}(\phi, T_z, \psi)$, $\cL_{\mathrm{bw}}(\phi, T_z, \psi)$, $\mathcal{L}_{\text{\algname}}(\phi, T_{z}, \psi)$, $\mathcal{L}_{\text{\algname}}(\psi, T_{z}, \phi)$. Under the same assumptions as Th.~\ref{th:mc-sp-vs-recon},
    \begin{enumerate}
        \item for all $z$, $\phi T_{\phi,z}^\star = \phi T_{z, \mathrm{fw}}^\star = \tilde\Pi_{\phi, z} M^{\pi_z} \psi$ and $\psi T_{\psi,z}^\star = \psi T_{z, \mathrm{bw}}^\star = \tilde\Pi_{\psi, z} M^{\pi_z} \phi$, where $\tilde\Pi_{\phi, z}$ (resp. $\tilde\Pi_{\psi, z}$) is an oblique projection on the span of $\phi$ (resp. $\psi$);
        \item $\nabla_{\phi} \mathcal{L}_{\text{\algname}}(\phi, T_z, \psi) = \nabla_{\phi} \cL_{\mathrm{fw}}(\phi, T_z, \psi)$ and $\nabla_{\psi} \mathcal{L}_{\text{\algname}}(\psi, T_z, \phi) = \nabla_{\psi} \cL_{\mathrm{fw}}(\phi, T_z, \psi)$.
    \end{enumerate}
\end{theorem}
\end{metaframe}
Similar to Th.~\ref{th:mc-sp-vs-recon}, the optimal predictors and gradients of \algname match those of the non-latent-predictive TD losses of Eq.~\ref{eq:td-den-fw} and \ref{eq:td-den-bw}, which are known to recover an approximation of the successor measure for bilinear parameterizations of the form $F_z^\transp B$ \citep{blier2021learning}. Unlike in the Monte Carlo case, here the optimal predictors solve a least-squares TD problem \citep{boyan1999least,precup2001off}, yielding the fixed point of a projected Bellman operator whose closed-form expression is an oblique projection \citep{scherrer2010should}.

\subsection{Policy evaluation and zero-shot RL} Finally, the following result motivates the significance of optimizing the successor measure losses of Eq.~\ref{eq:lr-reconstruction}, \ref{eq:td-den-fw}, and \ref{eq:td-den-bw}.
\begin{metaframe}
\begin{theorem}\label{th:policy-eval-bellman}
    Let $\phi, \psi$ have identity covariance matrices. For any reward function $r$, let $\omega_r := (\psi^\transp \psi)^{-1} \psi^\transp r$ be the linear regression weight for representation $\psi$. Then, for any $T_z$,
    \begin{align*}
        \max_{r \in \bR^{S} : \|r\|_2 \leq 1} \bE_{z\in\cZ} \left[ \sum_{s\in\cS}\left( V_{r}^{\pi_z}(s) - \phi(s)^\transp T_z \omega_r \right)^2 \right] \leq 2 \cL_{\mathrm{SM}}(\phi, T_z, \psi).
    \end{align*}
    Moreover, $\cL_{\mathrm{SM}}(\phi, T_z, \psi) \leq c \cL_{\mathrm{fw}}(\phi, T_z, \psi)$ and $\cL_{\mathrm{SM}}(\phi, T_z, \psi) \leq c \cL_{\mathrm{bw}}(\phi, T_z, \psi)$ for some $c$.
\end{theorem}
\end{metaframe}
Paraphrasing, the policy evaluation error of the technique in Section 3.3 (i.e., embed $r$ into a vector $\omega$ through linear regression on $\psi$, and compute $T_\phi(\phi(s),z)^\transp \omega$) is bounded by the successor measure approximation loss and the corresponding TD errors. Both these quantities are indirectly optimized by \algname (Th.~\ref{th:mc-sp-vs-recon}, \ref{th:td-sp-vs-den-new}), which is thus a sound approach for zero-shot policy evaluation. Moreover, Th.~\ref{th:policy-eval-bellman} leads to a zero-shot optimality result analogous to Theorem 2 of \citep{touati2021learning}: if the approximation of $M^{\pi_z}$ is perfect (i.e., $M^{\pi_z} = \phi T_z \psi^\transp$ for all $z$ or, equivalently, the TD errors in Eq.~\ref{eq:td-den-fw} and \ref{eq:td-den-bw} are zero) and the policies $\pi_z$ are optimal for all linear rewards in $\psi$, then the inference procedure above recovers optimal policies for \emph{any} (even non-linear) reward function.

\section{Experiments}
\label{sec:exp}

We benchmark zero-shot performance across a diverse set of problems, including 4 locomotion/navigation domains from ExoRL/DMC \citep{tassa2018deepmind, Yarats2022exorl}, as well as 9 navigation/manipulation domains from OGBench \citep{Park25ogbench}. The former suite involves reward-based tasks and high-coverage data, while the latter evaluates goal-reaching and provides low-coverage datasets\footnote{We additionally apply BC regularization in OGBench based on \citet{park2025flow}, as detailed in App.~\ref{app:impl.bc}}. 
We consider both proprioceptive and pixel-based variants of all domains, and report expected returns/success rates across a set of tasks (4-8 depending on the domain) as main evaluation metric. In DMC, we often normalize returns by the maximum achievable ($1000$).

We structure our evaluation in four parts: \textbf{(i)} a comprehensive evaluation of \algname with respect to existing zero-shot methods;
\textbf{(ii)} an ablation over the prediction target, 
measuring the impact of multi-step, policy-aware dynamics modeling;
\textbf{(iii)} a comparison of \algname to its symmetric variant that learns a shared state-task encoder $\phi$; and \textbf{(iv)} a demonstration of fast adaptation from pre-trained state representations. Further results are presented in App.~\ref{app:exp}, and implementation details in App.~\ref{app:impl}.

\subsection{How does \algname compare to zero-shot RL algorithms?}

\begin{table}
\vspace{-15pt}
\centering
\resizebox{\textwidth}{!}{
\begin{tabular}{lcccccccc}
\toprule
 & Laplacian & ICVF* & HILP & FB & RLDP & BYOL* & BYOL-$\gamma$* & TD-JEPA \\ \midrule
DMC$_{\text{RGB}}$ (\texttt{avg}) & 293.1 \scriptsize{$\pm$ 15.1} & 438.7 \scriptsize{$\pm$ 14.9} & 391.2 \scriptsize{$\pm$ 23.8} & 456.2 \scriptsize{$\pm$ 8.6} & 525.7 \scriptsize{$\pm$ 13.3} & 513.8 \scriptsize{$\pm$ 11.6} & 582.4 \scriptsize{$\pm$ 9.8} & \textbf{628.8 \scriptsize{$\pm$ 5.5}} \\
\quad \texttt{walker} & \scriptsize{309.4} \scriptsize{$\pm$ 50.0} & \scriptsize{534.9} \scriptsize{$\pm$ 61.3} & \scriptsize{422.8} \scriptsize{$\pm$ 32.5} & \scriptsize{324.4} \scriptsize{$\pm$ 16.6} & \scriptsize{576.1} \scriptsize{$\pm$ 35.3} & \scriptsize{595.2} \scriptsize{$\pm$ 9.0} & \scriptsize{648.3} \scriptsize{$\pm$ 36.5} & \textbf{\scriptsize{738.9} \scriptsize{$\pm$ 3.5}} \\
\quad \texttt{cheetah} & \scriptsize{242.4} \scriptsize{$\pm$ 29.6} & \scriptsize{394.9}  \scriptsize{$\pm$ 30.1} & \scriptsize{333.0} \scriptsize{$\pm$ 86.6} & \scriptsize{622.4} \scriptsize{$\pm$ 23.1} & \scriptsize{605.3} \scriptsize{$\pm$ 23.5} & \scriptsize{468.0} \scriptsize{$\pm$ 46.7} & \scriptsize{679.8} \scriptsize{$\pm$ 17.1} & \textbf{\scriptsize{706.0} \scriptsize{$\pm$ 4.1}} \\
\quad \texttt{quadruped} & \scriptsize{430.1} \scriptsize{$\pm$ 32.3} & \scriptsize{583.3} \scriptsize{$\pm$ 17.2} & \scriptsize{513.9} \scriptsize{$\pm$ 10.8} & \scriptsize{475.4} \scriptsize{$\pm$ 16.7} & \scriptsize{551.1} \scriptsize{$\pm$ 23.4} & \scriptsize{581.8} \scriptsize{$\pm$ 16.6} & \scriptsize{570.0} \scriptsize{$\pm$ 6.6} & \textbf{\scriptsize{626.7} \scriptsize{$\pm$ 13.6}} \\
\quad \texttt{pointmass} & \scriptsize{190.4} \scriptsize{$\pm$ 12.4} & \scriptsize{241.6} \scriptsize{$\pm$ 35.6} & \scriptsize{294.9} \scriptsize{$\pm$ 33.4} & \scriptsize{402.8} \scriptsize{$\pm$ 16.8} & \scriptsize{370.3} \scriptsize{$\pm$ 12.0} & \scriptsize{410.3} \scriptsize{$\pm$ 8.5} & \textbf{\scriptsize{431.6} \scriptsize{$\pm$ 17.4}} & \textbf{\scriptsize{443.7} \scriptsize{$\pm$ 10.9}} \\
\midrule
DMC (\texttt{avg}) & 591.1 \scriptsize{$\pm$ 10.7} & 619.3 \scriptsize{$\pm$ 10.3} & 620.1 \scriptsize{$\pm$ 8.4} & 648.2 \scriptsize{$\pm$ 4.1} & 610.2 \scriptsize{$\pm$ 13.5} & 618.6 \scriptsize{$\pm$ 10.5} & \textbf{645.4 \scriptsize{$\pm$ 10.5}} & \textbf{661.2 \scriptsize{$\pm$ 6.3}} \\
\quad \texttt{walker} & \scriptsize{769.7} \scriptsize{$\pm$ 4.7} & \scriptsize{727.0} \scriptsize{$\pm$ 16.2} & \scriptsize{796.4} \scriptsize{$\pm$ 7.7} & \textbf{\scriptsize{811.5} \scriptsize{$\pm$ 5.9}} & \scriptsize{723.9} \scriptsize{$\pm$ 18.3} & \scriptsize{746.8} \scriptsize{$\pm$ 11.0} & \scriptsize{786.1} \scriptsize{$\pm$ 9.6} & \scriptsize{785.2} \scriptsize{$\pm$ 6.7} \\
\quad \texttt{cheetah} & \scriptsize{614.5} \scriptsize{$\pm$ 18.9} & \scriptsize{606.3} \scriptsize{$\pm$ 16.8} & \scriptsize{618.3} \scriptsize{$\pm$ 5.8} & \scriptsize{672.7} \scriptsize{$\pm$ 4.9} & \scriptsize{575.6} \scriptsize{$\pm$ 44.9} & \scriptsize{622.8} \scriptsize{$\pm$ 23.9} & \scriptsize{647.2} \scriptsize{$\pm$ 9.0} & \textbf{\scriptsize{688.7} \scriptsize{$\pm$ 6.7}} \\
\quad \texttt{quadruped} & \scriptsize{635.0} \scriptsize{$\pm$ 38.7} & \textbf{\scriptsize{708.5} \scriptsize{$\pm$ 14.2}} & \textbf{\scriptsize{694.8} \scriptsize{$\pm$ 11.0}} & \scriptsize{595.6} \scriptsize{$\pm$ 9.1} & \scriptsize{665.0} \scriptsize{$\pm$ 13.9} & \scriptsize{611.8} \scriptsize{$\pm$ 28.1} & \textbf{\scriptsize{683.1} \scriptsize{$\pm$ 26.1}} & \textbf{\scriptsize{691.4} \scriptsize{$\pm$ 5.0}} \\
\quad \texttt{pointmass} & \scriptsize{345.1} \scriptsize{$\pm$ 22.4} & \scriptsize{435.5} \scriptsize{$\pm$ 11.1} & \scriptsize{371.0} \scriptsize{$\pm$ 37.1} & \textbf{\scriptsize{513.0} \scriptsize{$\pm$ 20.0}} & \textbf{\scriptsize{476.3} \scriptsize{$\pm$ 39.4}} & \textbf{\scriptsize{493.0} \scriptsize{$\pm$ 41.3}} & \scriptsize{465.1} \scriptsize{$\pm$ 17.6} & \textbf{\scriptsize{479.3} \scriptsize{$\pm$ 23.6}} \\
\midrule
OGBench$_{\text{RGB}}$ (\texttt{avg}) & 30.58 \scriptsize{$\pm$ 0.81} & 25.22 \scriptsize{$\pm$ 0.55} & 32.56 \scriptsize{$\pm$ 0.92} & 39.89 \scriptsize{$\pm$ 0.47} & 39.09 \scriptsize{$\pm$ 0.59} & 40.33 \scriptsize{$\pm$ 0.52} & \textbf{41.58 \scriptsize{$\pm$ 0.64}} & \textbf{41.34 \scriptsize{$\pm$ 0.45}} \\
\quad \texttt{antmaze-mn} & \scriptsize{92.20} \scriptsize{$\pm$ 2.91} & \scriptsize{85.80} \scriptsize{$\pm$ 3.02} & \scriptsize{84.60} \scriptsize{$\pm$ 3.59} & \textbf{\scriptsize{96.80} \scriptsize{$\pm$ 0.74}} & \textbf{\scriptsize{97.60} \scriptsize{$\pm$ 0.50}} & \scriptsize{94.40} \scriptsize{$\pm$ 1.48} & \textbf{\scriptsize{98.00} \scriptsize{$\pm$ 0.73}} & \textbf{\scriptsize{96.67} \scriptsize{$\pm$ 1.11}} \\
\quad \texttt{antmaze-ln} & \scriptsize{35.40} \scriptsize{$\pm$ 2.97} & \scriptsize{42.60} \scriptsize{$\pm$ 2.84} & \scriptsize{47.00} \scriptsize{$\pm$ 4.04} & \textbf{\scriptsize{76.80} \scriptsize{$\pm$ 2.33}} & \scriptsize{63.60} \scriptsize{$\pm$ 3.89} & \scriptsize{62.20} \scriptsize{$\pm$ 3.42} & \scriptsize{68.80} \scriptsize{$\pm$ 2.70} & \textbf{\scriptsize{74.60} \scriptsize{$\pm$ 3.35}} \\
\quad \texttt{antmaze-ms} & \scriptsize{60.20} \scriptsize{$\pm$ 3.88} & \scriptsize{46.20} \scriptsize{$\pm$ 2.74} & \scriptsize{71.80} \scriptsize{$\pm$ 2.22} & \scriptsize{86.20} \scriptsize{$\pm$ 2.05} & \textbf{\scriptsize{90.60} \scriptsize{$\pm$ 1.91}} & \textbf{\scriptsize{90.40} \scriptsize{$\pm$ 1.97}} & \textbf{\scriptsize{86.00} \scriptsize{$\pm$ 3.10}} & \scriptsize{84.40} \scriptsize{$\pm$ 3.85} \\
\quad \texttt{antmaze-ls} & \scriptsize{7.20} \scriptsize{$\pm$ 1.98} & \scriptsize{7.20} \scriptsize{$\pm$ 1.20} & \scriptsize{23.60} \scriptsize{$\pm$ 1.83} & \textbf{\scriptsize{27.40} \scriptsize{$\pm$ 2.78}} & \scriptsize{21.80} \scriptsize{$\pm$ 1.01} & \textbf{\scriptsize{26.60} \scriptsize{$\pm$ 2.23}} & \textbf{\scriptsize{28.60} \scriptsize{$\pm$ 1.71}} & \textbf{\scriptsize{28.80} \scriptsize{$\pm$ 2.50}} \\
\quad \texttt{antmaze-me} & \scriptsize{0.00} \scriptsize{$\pm$ 0.00} & \scriptsize{0.00} \scriptsize{$\pm$ 0.00} & \scriptsize{0.20} \scriptsize{$\pm$ 0.20} & \textbf{\scriptsize{1.80} \scriptsize{$\pm$ 1.09}} & \textbf{\scriptsize{0.80} \scriptsize{$\pm$ 0.44}} & \textbf{\scriptsize{1.20} \scriptsize{$\pm$ 1.00}} & \textbf{\scriptsize{3.20} \scriptsize{$\pm$ 1.98}} & \scriptsize{0.20} \scriptsize{$\pm$ 0.20} \\
\quad \texttt{cube-single} & \textbf{\scriptsize{73.80} \scriptsize{$\pm$ 3.53}} & \scriptsize{34.80} \scriptsize{$\pm$ 7.03} & \scriptsize{56.40} \scriptsize{$\pm$ 3.82} & \scriptsize{62.00} \scriptsize{$\pm$ 2.27} & \scriptsize{63.20} \scriptsize{$\pm$ 3.91} & \textbf{\scriptsize{75.40} \scriptsize{$\pm$ 2.58}} & \textbf{\scriptsize{76.40} \scriptsize{$\pm$ 3.24}} & \scriptsize{67.80} \scriptsize{$\pm$ 3.67} \\
\quad \texttt{cube-double} & \textbf{\scriptsize{1.60} \scriptsize{$\pm$ 0.72}} & \scriptsize{0.80} \scriptsize{$\pm$ 0.44} & \textbf{\scriptsize{1.60} \scriptsize{$\pm$ 0.58}} & \scriptsize{1.20} \scriptsize{$\pm$ 0.61} & \textbf{\scriptsize{2.20} \scriptsize{$\pm$ 1.31}} & \textbf{\scriptsize{2.40} \scriptsize{$\pm$ 0.65}} & \scriptsize{1.40} \scriptsize{$\pm$ 0.67} & \textbf{\scriptsize{3.00} \scriptsize{$\pm$ 0.91}} \\
\quad \texttt{scene} & \scriptsize{2.80} \scriptsize{$\pm$ 1.12} & \scriptsize{8.40} \scriptsize{$\pm$ 1.45} & \scriptsize{5.40} \scriptsize{$\pm$ 1.63} & \scriptsize{4.20} \scriptsize{$\pm$ 0.87} & \scriptsize{9.40} \scriptsize{$\pm$ 1.33} & \scriptsize{8.80} \scriptsize{$\pm$ 1.64} & \textbf{\scriptsize{11.20} \scriptsize{$\pm$ 1.82}} & \textbf{\scriptsize{14.20} \scriptsize{$\pm$ 2.22}} \\
\quad \texttt{puzzle-3x3} & \textbf{\scriptsize{2.00} \scriptsize{$\pm$ 1.40}} & \scriptsize{1.20} \scriptsize{$\pm$ 0.44} & \textbf{\scriptsize{2.44} \scriptsize{$\pm$ 0.99}} & \textbf{\scriptsize{2.60} \scriptsize{$\pm$ 0.79}} & \textbf{\scriptsize{2.60} \scriptsize{$\pm$ 0.79}} & \textbf{\scriptsize{1.60} \scriptsize{$\pm$ 0.40}} & \scriptsize{0.60} \scriptsize{$\pm$ 0.31} & \textbf{\scriptsize{2.40} \scriptsize{$\pm$ 0.83}} \\
\midrule
OGBench (\texttt{avg}) & 14.81 \scriptsize{$\pm$ 1.32} & 30.87 \scriptsize{$\pm$ 0.58} & \textbf{37.98 \scriptsize{$\pm$ 1.11}} & \textbf{39.04 \scriptsize{$\pm$ 0.66}} & 27.07 \scriptsize{$\pm$ 0.83} & 26.42 \scriptsize{$\pm$ 0.83} & 30.42 \scriptsize{$\pm$ 0.94} & \textbf{37.98 \scriptsize{$\pm$ 0.77}} \\
\quad \texttt{antmaze-mn} & \scriptsize{50.00} \scriptsize{$\pm$ 4.94} & \textbf{\scriptsize{79.80} \scriptsize{$\pm$ 2.62}} & \textbf{\scriptsize{83.60} \scriptsize{$\pm$ 2.63}} & \scriptsize{73.00} \scriptsize{$\pm$ 2.72} & \scriptsize{74.60} \scriptsize{$\pm$ 4.15} & \scriptsize{58.40} \scriptsize{$\pm$ 2.00} & \scriptsize{51.40} \scriptsize{$\pm$ 1.55} & \scriptsize{70.40} \scriptsize{$\pm$ 3.72} \\
\quad \texttt{antmaze-ln} & \scriptsize{21.60} \scriptsize{$\pm$ 3.90} & \textbf{\scriptsize{58.40} \scriptsize{$\pm$ 1.90}} & \scriptsize{52.60} \scriptsize{$\pm$ 3.86} & \scriptsize{36.80} \scriptsize{$\pm$ 4.28} & \scriptsize{36.40} \scriptsize{$\pm$ 4.66} & \scriptsize{26.60} \scriptsize{$\pm$ 3.03} & \scriptsize{21.80} \scriptsize{$\pm$ 3.57} & \textbf{\scriptsize{57.20} \scriptsize{$\pm$ 4.25}} \\
\quad \texttt{antmaze-ms} & \scriptsize{21.40} \scriptsize{$\pm$ 4.32} & \scriptsize{39.00} \scriptsize{$\pm$ 3.30} & \scriptsize{50.60} \scriptsize{$\pm$ 2.46} & \textbf{\scriptsize{70.40} \scriptsize{$\pm$ 3.95}} & \scriptsize{58.40} \scriptsize{$\pm$ 3.29} & \scriptsize{60.60} \scriptsize{$\pm$ 5.07} & \scriptsize{45.60} \scriptsize{$\pm$ 2.84} & \scriptsize{61.56} \scriptsize{$\pm$ 4.53} \\
\quad \texttt{antmaze-ls} & \scriptsize{11.80} \scriptsize{$\pm$ 1.47} & \scriptsize{13.20} \scriptsize{$\pm$ 1.64} & \scriptsize{12.20} \scriptsize{$\pm$ 1.75} & \textbf{\scriptsize{49.80} \scriptsize{$\pm$ 5.64}} & \scriptsize{19.60} \scriptsize{$\pm$ 2.73} & \scriptsize{25.80} \scriptsize{$\pm$ 4.28} & \scriptsize{20.20} \scriptsize{$\pm$ 1.80} & \scriptsize{40.60} \scriptsize{$\pm$ 2.51} \\
\quad \texttt{antmaze-me} & \scriptsize{0.80} \scriptsize{$\pm$ 0.61} & \scriptsize{0.00} \scriptsize{$\pm$ 0.00} & \scriptsize{2.00} \scriptsize{$\pm$ 0.84} & \textbf{\scriptsize{51.60} \scriptsize{$\pm$ 2.65}} & \scriptsize{4.80} \scriptsize{$\pm$ 2.35} & \scriptsize{11.40} \scriptsize{$\pm$ 2.29} & \scriptsize{19.60} \scriptsize{$\pm$ 2.53} & \scriptsize{20.20} \scriptsize{$\pm$ 2.39} \\
\quad \texttt{cube-single} & \scriptsize{15.11} \scriptsize{$\pm$ 1.49} & \scriptsize{20.40} \scriptsize{$\pm$ 1.93} & \textbf{\scriptsize{74.20} \scriptsize{$\pm$ 3.53}} & \scriptsize{49.60} \scriptsize{$\pm$ 3.83} & \scriptsize{19.80} \scriptsize{$\pm$ 2.41} & \scriptsize{22.00} \scriptsize{$\pm$ 3.16} & \textbf{\scriptsize{79.40} \scriptsize{$\pm$ 2.83}} & \scriptsize{34.20} \scriptsize{$\pm$ 2.88} \\
\quad \texttt{cube-double} & \scriptsize{2.00} \scriptsize{$\pm$ 0.42} & \scriptsize{5.00} \scriptsize{$\pm$ 0.80} & \textbf{\scriptsize{20.00} \scriptsize{$\pm$ 2.72}} & \scriptsize{2.60} \scriptsize{$\pm$ 0.43} & \scriptsize{3.80} \scriptsize{$\pm$ 0.76} & \scriptsize{4.40} \scriptsize{$\pm$ 0.72} & \scriptsize{2.60} \scriptsize{$\pm$ 0.67} & \scriptsize{3.60} \scriptsize{$\pm$ 0.78} \\
\quad \texttt{scene} & \scriptsize{7.80} \scriptsize{$\pm$ 1.28} & \textbf{\scriptsize{45.40} \scriptsize{$\pm$ 2.29}} & \textbf{\scriptsize{43.80} \scriptsize{$\pm$ 1.90}} & \scriptsize{12.80} \scriptsize{$\pm$ 1.61} & \scriptsize{11.60} \scriptsize{$\pm$ 1.57} & \scriptsize{15.40} \scriptsize{$\pm$ 1.37} & \scriptsize{14.40} \scriptsize{$\pm$ 2.32} & \scriptsize{38.44} \scriptsize{$\pm$ 1.37} \\
\quad \texttt{puzzle-3x3} & \scriptsize{2.80} \scriptsize{$\pm$ 0.68} & \scriptsize{16.60} \scriptsize{$\pm$ 0.73} & \scriptsize{2.80} \scriptsize{$\pm$ 0.68} & \scriptsize{4.80} \scriptsize{$\pm$ 0.68} & \scriptsize{14.60} \scriptsize{$\pm$ 0.90} & \scriptsize{13.20} \scriptsize{$\pm$ 1.91} & \textbf{\scriptsize{18.80} \scriptsize{$\pm$ 0.44}} & \scriptsize{15.60} \scriptsize{$\pm$ 1.11} \\
\bottomrule
\end{tabular}
}
\vspace{-8pt}
\caption{Performance of zero-shot algorithms for DMC (reward) and OGBench (success rate) with either proprioception or RGB inputs. We report means and standard errors across seeds. Numbers are bold for top algorithms if confidence intervals overlap.}
\label{tab:main.results}
\vspace{-10pt}
\end{table}

\begin{figure}[t]
    \centering
    \includegraphics[width=0.9\textwidth]{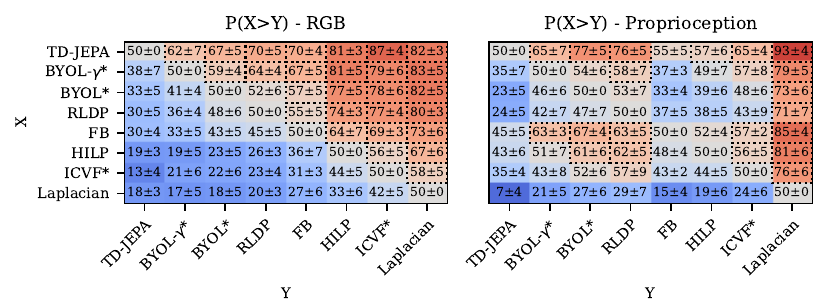}
    \caption{Probabilities of improvement: how lixely is method X to outperform method Y on a random domain? We report symmetrized 95\% simple bootstrap confidence intervals. Dotted lines surround matches in which the improvement is statistically significant.}
    \label{fig:wr}
\end{figure}

We first compare \algname to three groups of successor-feature-based zero-shot RL baselines:\footnote{Notice that only \textit{Laplacian}, \textit{HILP}, \textit{FB} and \textit{RLDP} are standard zero-shot unsupervised RL algorithms, while \textit{BYOL}, \textit{BYOL-$\gamma$}, and \textit{ICVF} (henceforth marked with a $*$) are representation learning methods: their instantiation in a zero-shot framework is novel and designed to investigate the impact of different representations.}
\begin{itemize}
    \item \textit{Laplacian} \citep{wu2018laplacian}, \textit{HILP} \citep{park2024foundation}, and \textit{FB} \citep{touati2021learning} are established zero-shot methods that train a task encoder $\psi$, without specific learning objectives for a state encoder.
    \item \textit{BYOL$^\star$} \citep{grill2020bootstrap}, \textit{BYOL-$\gamma^\star$} \citep{lawson2025self} and \textit{RLDP} \citep{jajoo2025regularized} learn a state encoder $\phi$ via latent-predictive learning, which we then use as a task encoder for successor features (learned through a contrastive loss in the case of RLDP).
    \item \textit{ICVF$^\star$} \citep{ghosh2023reinforcement} learns a multilinear decomposition of the successor measure via expectile regression, yielding both state and task encoders on top of which we train successor features.
\end{itemize} 
For a fair comparison, each method is tuned over comparable hyperparameter grids and adopts the same architecture:
in particular, the state input is always passed through an explicit state encoder before being fed into, e.g., the successor features estimator $F(s,a;z)$\footnote{On average, explicit state encoders actually improve the performance for existing methods, see App.~\ref{app:exp.state_encoder}.}.
We find that this protocol results in significant improvements in zero-shot performances, even for existing methods (e.g., $1.3\times$ and $2.4\times$ higher than overlapping pixel-based results for the methods presented in \cite{park2024foundation} and \cite{jajoo2025regularized}, respectively), as displayed in Tab~\ref{tab:main.results}.
When considering suite-aggregated performance, we find that \algname is on par or better than the best performing baseline in each suite.
Given the diverse nature of suites (proprioception vs pixels), domains (locomotion, navigation, manipulation) and datasets (high- vs low-coverage), many algorithms unsurprisingly achieve strong performance in some configurations while under-performing in others.
We thus additionally measure how consistently well each algorithm performs by computing the probability of improvement \citep{agarwal2021deep} across all domains in Fig.~\ref{fig:wr}.
We find that \algname is consistently among the top performing algorithms, whereas most baselines perform well on a narrow subset of problems. For instance, while \algname is only slightly preferable to FB and HILP from proprioception, it is significantly better than them in visual domains. Similarly, BYOL-$\gamma$ is slightly better than \algname in OGBench$_\textrm{RGB}$, but it is significantly worse in DMC$_\textrm{RGB}$ and OGBench. Finally, we note that latent-predictive methods tend to be generally preferrable in pixel-based domains.

\subsection{Which dynamics should latent-predictive zero-shot algorithms model?}

\begin{figure}
    \centering
    \begin{minipage}{.5\textwidth}
    \centering
    \includegraphics[width=\textwidth]{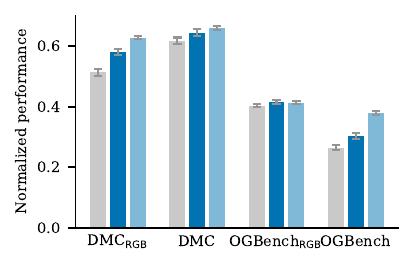}
    \label{fig:spr}
    \begin{minipage}{\textwidth}
    \hspace{4em}
    { 
    \footnotesize
    \raisebox{0.4ex}{\textcolor{gray}{\rule{10pt}{2pt}}} BYOL* \;
    \raisebox{0.4ex}{\textcolor{our_blue}{\rule{10pt}{2pt}}} BYOL-$\gamma$* \;
    \raisebox{0.4ex}{\textcolor{our_bluel}{\rule{10pt}{2pt}}} \algname \;
    }
    \end{minipage}
    \end{minipage}%
    \begin{minipage}{0.5\textwidth}
    \centering
    \includegraphics[width=0.9\textwidth]{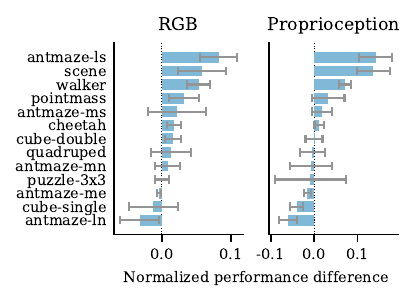}
    \label{fig:zs_sptdff_sptd}
    \end{minipage}%
    \caption{\textbf{Left:} normalized zero-shot performance for latent-predictive methods. \textbf{Right:} difference in normalized performance between \algname and its symmetric variant. Error bars represent standard errors on normalized performance or its differences, respectively.}
    \label{fig:sptd-vs-spr-vs-symm}
\end{figure}

The baselines based on BYOL and BYOL-$\gamma$ are algorithmically closest to \algname, and allow a precise investigation on the dynamics to model. While BYOL$^\star$ and BYOL-$\gamma^\star$ approximate one-step and multi-step transitions of the behavioral policy, respectively, \algname models multi-step transitions \emph{of the zero-shot policies}. While approximating the behavioral dynamics can be effective for expert-like data (i.e., in OGBench), we observe a general pattern suggesting that directly modeling policy-conditional successor measures is on average beneficial, as reported in Fig.~\ref{fig:sptd-vs-spr-vs-symm} (\emph{left}).

\subsection{Should state and task representations differ?}

\algname trains separate state and task encoders: while this may grant a better approximation of successor measures, sharing state and task representations while optimizing a single objective (see Section~\ref{sec:sptd-symm}) may in practice be more efficient. We measure the difference in per-task normalized performance between \algname and a symmetric variant in Figure~\ref{fig:sptd-vs-spr-vs-symm} (\emph{right}): we observe that this variant performs comparatively rather well, while relying on a single predictor-encoder pair. However, using distinct state and task embeddings tends to improve empirical performance more often than not. \looseness -1

\subsection{Are state representations beneficial for fast adaptation?}

\begin{figure}
    \centering
    \begin{minipage}{0.9\textwidth}
    \includegraphics[width=\textwidth]{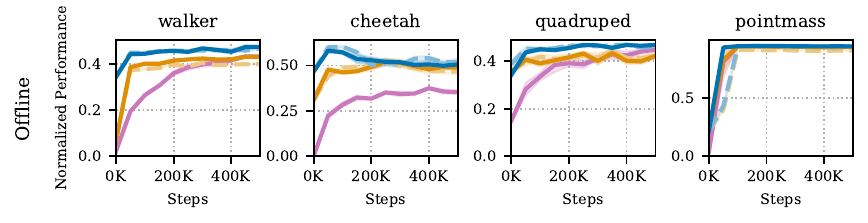}
    \end{minipage}\\
    \begin{minipage}{0.9\textwidth}
    \includegraphics[width=\textwidth]{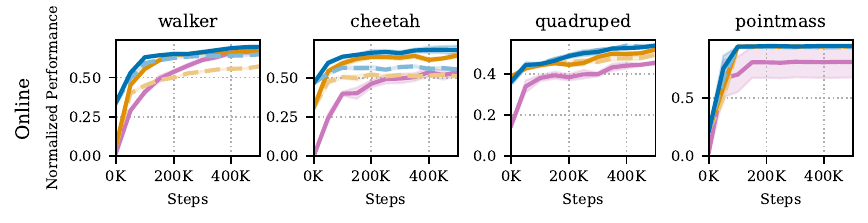}
    \end{minipage}\\
    {
    \footnotesize
    \centering
    \raisebox{0.4ex}{\textcolor{our_blue}{\rule{10pt}{2pt}}} \; \algname \quad
    \raisebox{0.4ex}{\textcolor{our_bluel}{\rule{4pt}{2pt}\,\rule{4pt}{2pt}}} \; \algname (frozen) \quad
    \raisebox{0.4ex}{\textcolor{our_orange}{\rule{10pt}{2pt}}} \; FB \quad
    \raisebox{0.4ex}{\textcolor{our_orangel}{\rule{4pt}{2pt}\,\rule{4pt}{2pt}}} \; FB (frozen) \quad
    \raisebox{0.4ex}{\textcolor{our_pink}{\rule{10pt}{2pt}}} \; Scratch \quad
    }
    \caption{Normalized performance of zero-shot policies when fine-tuned offline (top) or online (bottom). Initializing the agent to zero-shot solutions (blue and yellow lines) results in sample-efficient learning; frozen representations (dashed) are often expressive enough to enable fast adaptation.}
    \label{fig:fast-adapt}
\end{figure}

While the previous evaluations have focused on  aggregated zero-shot performance, we now investigate an additional benefit of explicit state representations: fast adaptation at test-time.
Given a pixel-based task, we initialize the agent with the zero-shot policy $\pi_{z}$ and critic learned at pre-training, and we either \emph{fine-tune} the whole model via TD3 \citep{fujimoto2018addressing} or keep the pre-trained state encoder \emph{frozen}. We consider two RL adaptation protocols (i) \textbf{Offline}: a transition-reward dataset is provided $\mathcal{D}_{\text{rew}} = \{ (s,a,s',r) \}$ and TD3 updates are applied offline; (ii) \textbf{Online}: an online buffer is additionally collected over time and batches are sampled by mixing it with the offline buffer mentioned above (following the unsupervised-to-online protocol of \cite{kim2024unsupervised}). 
Figure \ref{fig:fast-adapt} reports results for each DMC domain for the task in which the gap between online and zero-shot algorithms is largest; we consider \algname and FB as strong, representative algorithms among self-predictive and contrastive methods.
We first observe that fine-tuning pre-trained agents leads to large gains in sample efficiency w.r.t. training from scratch, and reaches the asymptotic performance of TD3.
More interestingly, frozen representations are often sufficient for downstream learning, and do not need further fine-tuning. We refer to App.~\ref{app:exp.adapt} and App.~\ref{app:impl.adapt} for further results and details, respectively.

\section{Conclusion}

Through the introduction of a novel temporal-difference latent-predictive loss, we presented a zero-shot unsupervised RL method that operates entirely in latent space and can be shown to recover a factorization of the successor measures of multiple policies.
Empirically, we found that \algname matches the best zero-shot methods when learning from proprioception, and exceeds them when learning from pixels, while also retrieving state representations that allow fast downstream adaptation.
As formal guarantees rely on an assumption of symmetry, one exciting direction for future work may study learning objectives that are compatible with asymmetric successor measures, yet remain amenable to practical optimization.
On a practical note, we believe that benchmarking latent-predictive zero-shot objectives on large-scale, real robotic dataset can shed further light on opportunities and limitations of this promising framework.

\clearpage
\newpage
\bibliographystyle{assets/plainnat}
\bibliography{bibliography}

\clearpage
\newpage
\beginappendix

\startcontents[sections]
\printcontents[sections]{l}{1}{\setcounter{tocdepth}{2}}

\clearpage

\section{Extended Related Work}

As \algname bridges zero-shot reinforcement learning and latent-predictive representation learning, we reserve this section to building connections to several related works in both areas beyond what was discussed in Sec.~\ref{sec:intro}.

\paragraph{Zero-shot RL algorithms}

These methods aim at pre-training agents on unsupervised data to enable solving a wide range of downstream tasks specified via reward functions in a zero-shot fashion, i.e., without additional test-time learning or planning. So far they have achieved impressive results, yielding so-called behavioral foundation models \citep{pirotta2024fast,tirinzoni2025zeroshot}. The forward-backward algorithm (FB, \citet{touati2021learning,touati2023does}) is an established method in this class, and perhaps the most related to \algname. FB learns a task encoder and estimates its successor features, essentially finding a bilinear decomposition of policy-conditional successor measures (e.g., $M^{\pi_z} \approx F_z B^\transp$). On the other hand, \algname uses the parameterization $M^{\pi_z} \approx \phi T_z \psi^\transp$, which thus explicitly trains shared (across tasks) state representations and enforces more structure in the predictor. On top of the difference in parameterization, FB adopts a \textit{contrastive} loss, which computes pairwise dot products across each training batch. This is not necessary for the objective of \algname, which is latent-predictive at its core. FB has further been shown capable of zero-shot imitation \citep{pirotta2024fast} and extended to several settings, including online training regularized by expert data \citep{tirinzoni2025zeroshot}, offline training on low-quality data \citep{jeen2024zeroshot}, training on environments with different dynamics \citep{bobrin2025zero}, online fine-tuning \citep{sikchi2025fast} and pure exploration \citep{urpi2025epistemically}.

Other methods, like HILP, PSM, and RLDP, can also be seen as training a task encoder $\psi$ plus successor features on top. HILP \citep{park2024foundation} trains $\psi$ through a ``goal-reaching'' loss that preserves temporal distance from true to latent state space. PSM \citep{agarwal2024proto} does so by learning an affine decomposition of the successor measure for a discrete
codebook of policies. Finally, RLDP \citep{jajoo2025regularized} trains $\phi$ using a chained multi-step latent-predictive loss similar to the one used by TD-MPC \citep{hansen2023td}. \cite{jajoo2025regularized} also observe that regularizing the representation to be orthonormal is crucial to avoid collapse and obtain good performance, which we also observe in \algname (see Alg.~\ref{alg:sptd}).

\paragraph{Latent-predictive methods}

As discussed in Section \ref{sec:intro}, these methods have mostly been applied to define auxiliary losses for a variety of RL settings. \cite{schwarzer2020data} use a latent-predictive loss to enhance state representations learned through a deep Q network. \cite{guo2020bootstrap} use latent prediction in the context of POMDPs to embed observations and histories. Their method trains two representations to be self-predictive of each other, hence connecting to the asymmetric variant of TD-JEPA and the method explained in Appendix \ref{app:theory-general}. \cite{hansen2023td} uses latent prediction to train a latent dynamics model that, combined with TD3, enables mixing model-free and model-based reinforcement learning, e.g., by test time planning to improve the pre-trained policy. \cite{sobal2025learning}, on the other hand, learn latent dynamics models on purely unsupervised data and show they can solve goal-reaching tasks via test-time planning.

BYOL-$\gamma$ \citep{lawson2025self} is a recent, closely related method for representation learning. At its core, BYOL-$\gamma$ predicts discounted future representations of states visited by the behavior policy, while \algname extends this objective to policy-conditional prediction. As a result, BYOL-$\gamma$ may be seen as an unconditional, Monte Carlo version of \algname, with a strict requirement in terms of on-policy data.
On the other hand, the on-policy nature of the algorithm enables \citet{lawson2025self} to implement a bi-directional update of asymmetric representations. \algname can also recover an asymmetric parameterization, but its practical objective is not bi-directional (i.e., it only implements forward TD prediction, cf. Appendix \ref{app:forward-backward} for a formal definition of the bi-directional latent-predictive objective).
Crucially, BYOL-$\gamma$ is not proposed as a zero-shot method: the version evaluated in this work is a novel instantiation in a successor feature framework.

\paragraph{Theory of latent-predictive representations}

The theory of latent-predictive or self-predictive representations has been previously studied in several works \citep{tang2023understanding, voelcker2024does,khetarpal2024unifying, lawson2025self}, with a particular focus on single-policy, single-step prediction (potentially, bi-directional). Our analysis of MC-JEPA (Section \ref{sec:theory}) largely takes place in a multi-policy setting, with generic transition kernels over states; as such, it subsumes and expands on several existing results (see Appendix \ref{app:generalize-existing}). On the other hand, representation learning through temporal difference losses, as in \algname, is largely understudied. The closest studies are by \cite{blier2021learning} and \cite{lan2023bootstrapped} which show that, under certain parameterizations and assumptions, TD representation learning can recover low-rank decompositions of the successor measure, in the sense that they optimize the corresponding approximation loss. These works crucially rely on having a single policy, and it remains an open questions whether such results extend to multiple policies. In this direction, we provide a first result which connects latent-predictive TD learning with TD learning over the successor measure for multiple polies.

\paragraph{Other representation learning methods for RL}

Beyond latent-predictive methods, ICVF \citep{ghosh2023reinforcement,bhateja2023robotic} has also been proposed as a multi-policy, multi-step representation learning objective. By relying on an implicit, value-based loss, this method may be applied to off-policy, action-free, transition-based data to recover a decomposition of policy-dependent successor measures. This decomposition is multi-linear, but, unlike \algname, it is restricted to linear predictors; its practical implementation moreover forces the same dimensionality across state and task embeddings.
As BYOL-$\gamma$, ICVF does not natively support zero-shot RL: we thus integrate it into a successor-features-based zero-shot policy optimization scheme. 

VIP \citep{ma2022vip} also works in a similar setting as ICVF. It casts representation learning as an offline goal-conditioned reinforcement learning problem. This can be seen related to approximating the successor measure of several goal-reaching policies. On the other hand, \algname does so for all reward-maximing policies, hence encompassing the goal-reaching ones.

MR.Q \citep{fujimoto2025towards} learns model-based representations that approximately linearize value functions for reward-based RL. This is achieved by combining reward prediction with a single-step latent dynamics loss similar to the one of BYOL, yielding state encoders on top of which value functions and policies are trained. \algname also aims at linearizing value functions, but does so with a multi-step policy-dependent loss and, most importantly, in a reward-free manner.

More broadly, several recent works evaluate visual representations pre-trained from large-scale data (e.g., internet videos) in control problems \citep{parisi2022unsurprising,majumdar2023we,silwal2024we,schneider2024surprising,zhou2024dino,mccarthy2025towards,tsagkas2025pre}. While there is no ``best'' method overall, as performance are problem and data dependent \citep{majumdar2023we}, representations pre-trained or fine-tuned with RL-related objectives, like time and task awareness \citep{ma2022vip,bhateja2023robotic,tsagkas2025pre}, perform well in general. \algname aligns with this view, as it shows that visual representations pre-trained with multi-step and policy-dependent objectives are suitable for value estimation and optimization across multiple tasks.

\section{Proofs}\label{app:proofs}

\subsection{Proof of Proposition \ref{prop:pred.succ.features}}

For any $z,s,a,s^+$, the term inside the expectation in Eq.~\ref{eq:mc-obj} can be rewritten as
\begin{align*}
    \| T_\phi(\phi(s),a,z) - \sg{\phi(s^+)}\|^2 &= \| T_\phi(\phi(s),a,z) \pm \sg{F_\phi^{\pi_z}(s,a)} - \sg{\phi(s^+)}\|^2
    \\ &= \| T_\phi(\phi(s),a,z) - \sg{F_\phi^{\pi_z}(s,a)}\|^2 + \| \sg{F_\phi^{\pi_z}(s,a)} - \sg{\phi(s^+)}\|^2
    \\ & \qquad\qquad - (T_\phi(\phi(s),a,z) - \sg{F_\phi^{\pi_z}(s,a)})^\transp (\sg{F_\phi^{\pi_z}(s,a)} - \sg{\phi(s^+)}).
\end{align*}
Taking the expectation w.r.t. $s^+\sim M^{\pi_z}(\cdot|s,a)$, it is easy to see that the last term is zero since $F_\phi^{\pi_z}(s,a) := \bE_{s^+\sim M^{\pi_z}(\cdot|s,a)}[\phi(s^+)]$, while the second term is a constant. This concludes the proof. 

\qedwhite

\begin{remark}
    By replacing the target $\sg{\phi(s^+)}$ above with $\sg{\psi(s^+)}$ and $F_\phi^{\pi_z}(s,a)$ with $F_\psi^{\pi_z}(s,a)$, this proof generalizes to the Monte Carlo loss with asymmetric representations (Eq.~\ref{eq:mc-fw-sp-loss}).
\end{remark}

\subsection{Proof of Theorem \ref{th:mc-sp-vs-recon}}

We begin by rewriting the MC-JEPA loss of Eq.~\ref{eq:mc-fw-sp-loss} with the notation of Sec.~\ref{sec:theory} as
\begin{align}\label{eq:sp-loss-matrix-form}
    \cL_{\text{MC-JEPA}}(\phi, T_{\phi,z}, \psi) := \frac{1}{2}\bE_{z\sim\cZ,s\sim\rho,s^+\sim M^{\pi_z}(\cdot | s)} \left[ \Big\| T_{\phi,z}^\transp \phi(s) - \sg{\psi(s^+)} \Big\|_2^2 \right],
\end{align}
where $\rho$ denotes the state distribution. Proposition \ref{prop:pred.succ.features} implies that $\cL_{\text{MC-JEPA}}$ has the same gradients as 
\begin{align*}
    \cL_{\mathrm{mc}}(\phi, T_{\phi,z}, \psi) := \frac{1}{2}\bE_{z\sim\cZ,s\sim\rho} \left[ \Big\| T_{\phi,z}^\transp \phi(s) - \bE_{s^+\sim M^{\pi_z}(\cdot | s)}\big[\sg{\psi(s^+)}\big] \Big\|_2^2 \right].
\end{align*}
Let $D_\rho \in \bR^{S\times S}$ be a diagonal matrix containing $\rho(s)$ for all states $s\in\cS$ on its diagonal. Using that $D_\rho = I$ by Assumption A2,\footnote{We ignore the $1/S$ scaling that only multiplies the loss by a constant.}
\begin{align}
    \cL_{\mathrm{mc}}(\phi, T_z, \psi) := \frac{1}{2}\bE_{z \sim \cZ}\left[ \| D_\rho^{1/2} (\phi T_z - \sg{M^{\pi_z}\psi}) \|_F^2 \right] = \frac{1}{2}\bE_{z \sim \cZ}\left[ \| \phi T_z - \sg{M^{\pi_z}\psi} \|_F^2 \right].
\end{align}
We now prove all statements for $\cL_{\mathrm{mc}}$, as gradient equivalence with Eq.~\ref{eq:sp-loss-matrix-form} implies they also hold for $\cL_{\text{MC-JEPA}}$.

\paragraph{Statement 1} 

Let us first compute the optimal predictors. For any $z\in\cZ$, the gradient of $\cL_{\mathrm{mc}}(\phi, T_z, \psi)$ w.r.t. $T_z$ is
\begin{align*}
    \nabla_{T_z} \cL_{\mathrm{mc}}(\phi, T_z, \psi) = p(z) \phi^\transp (\phi T_z - M^{\pi_z}\psi) = p(z) (T_z - \phi^\transp M^{\pi_z}\psi),
\end{align*}
where $p(z)$ is the probability to sample $z$\footnote{Without loss of generality, we also assume that $p(z) > 0$ for all $z$. If this is not the case, any $z$ with $p(z)=0$ can be removed from the loss.}, while the second equality uses that $\phi^\transp \phi = I$ by Assumption A1. This yields $T_{\phi,z}^\star = \phi^\transp M^{\pi_z}\psi$. Moreover, by simply inverting the roles of $\phi$ and $\psi$, we find that $\nabla_{T_z} \cL_{\mathrm{mc}}(\psi, T_z, \phi) = p(z)(T_z - \psi^\transp M^{\pi_z}\phi)$ and, thus, $T_{\psi,z}^\star = \psi^\transp M^{\pi_z}\phi$.

The gradient of $\cL_{\mathrm{SM}}(\phi, T_z, \psi)$ w.r.t. $T_z$ is
\begin{align*}
    \nabla_{T_z} \cL_{\mathrm{SM}}(\phi, T_z, \psi) = p(z)\phi^\transp (\phi T_z \psi^\transp -  M^{\pi_z})\psi = p(z)(T_z -  \phi^\transp  M^{\pi_z} \psi),
\end{align*}
where we used again Assumption A1. Hence, $T_z^\star = \phi^\transp M^{\pi_z}\psi$. Therefore, we clearly have that $T_z^\star = T_{\phi,z}^\star$. Moreover, since $P^{\pi_z}$ is symmetric by Assumption A3, $M^{\pi_z}$ is symmetric too, and 
$$T_{\psi,z}^\star = (\phi^\transp (M^{\pi_z})^\transp \psi)^\transp = (\phi^\transp M^{\pi_z} \psi)^\transp = (T_{z}^\star)^\transp.$$
Finally, it is easy to see that $\phi T_{\phi,z}^\star$ and $\psi T_{\psi,z}^\star$ satisfy the stated expressions for $\Pi_\phi := \phi \phi^\transp$ and $\Pi_\psi := \psi \psi^\transp$, respectively. Moreover, $\Pi_\phi$ and $\Pi_\psi$ are symmetric and idempotent ($\Pi_\phi \Pi_\phi = \Pi_\phi$), hence they are orthogonal projection matrices. This proves the first part of the statement.

\paragraph{Statement 2}

Let us now fix any $T_z$ for all $z\in\cZ$ and compute the gradients w.r.t. $\phi$ and $\psi$.
\begin{align*}
    \nabla_{\phi} \cL_{\mathrm{mc}}(\phi, T_z, \psi) &= \bE_{z\sim\cZ}\big[ (\phi T_z - M^{\pi_z}\psi) T_z^\transp \big],
    \\ \nabla_{\psi} \cL_{\mathrm{mc}}(\psi, T_z, \phi) &= \bE_{z\sim\cZ}\big[ (\psi T_z - M^{\pi_z}\phi) T_z^\transp \big],
    \\ \nabla_{\phi} \cL_{\mathrm{SM}}(\phi, T_z, \psi) &= \bE_{z\sim\cZ}\big[ (\phi T_z \psi^\transp - M^{\pi_z}) \psi T_z^\transp \big] = \bE_{z\sim\cZ}\big[ (\phi T_z - M^{\pi_z} \psi) T_z^\transp \big],
    \\ \nabla_{\psi} \cL_{\mathrm{SM}}(\phi, T_z^\transp, \psi) &= \bE_{z\sim\cZ}\big[ (\psi T_z \phi^\transp - (M^{\pi_z})^\transp) \phi T_z^\transp \big] = \bE_{z\sim\cZ}\big[ (\psi T_z - (M^{\pi_z})^\transp \phi) T_z^\transp \big],
\end{align*}
where we used Assumption A1 to simplify the last two expressions. Given that $P^{\pi_z}$ and, thus, $M^{\pi_z}$ are symmetric by Assumption A3, these gradients match as stated.

\qedwhite

\subsection{Proof of Theorem \ref{th:td-constant-cov}}\label{app:proof-non-collapse}

We begin by rewriting the TD-JEPA loss of Eq.~\ref{eq:asymm.sptd.phi} with the notation of Sec.~\ref{sec:theory} as
\begin{align}\label{eq:td-loss-matrix-form2}
    \mathcal{L}_{\text{\algname}}(\phi, T_z, \psi) := \frac{1}{2}\bE_{z\sim\cZ,s\sim\rho,s^+\sim P^{\pi_z}(\cdot | s)} \left[ \Big\| T_{\phi,z}^\transp \phi(s) - \sg{\psi(s^+)} - \gamma  \sg{T_{\phi,z}^\transp \phi(s^+)} \Big\|_2^2 \right].
\end{align} 
Following the proof of Theorem \ref{th:mc-sp-vs-recon}, we can put this in matrix form as
\begin{align}\label{eq:sp-td-matrix2}
    \cL_{\mathrm{td}}(\phi, T_z, \psi) := \frac{1}{2}\bE_{z\sim\cZ} \left[ \Big\| D_\rho^{1/2} \left( \phi T_{z} - U_z \right) \Big\|_F^2 \right],
\end{align}   
where $U_z := \sg{P^{\pi_z}\psi} - \gamma  \sg{P^{\pi_z} \phi T_{z}}$.
As we only brought expectations inside the norm, Eq.~\ref{eq:sp-td-matrix2} has the same gradients as Eq.~\ref{eq:td-loss-matrix-form2}. Hence, we define a continuous-time relaxation of gradient descend dynamics for Eq.~\ref{eq:sp-td-matrix2} (equiv. Eq.~\ref{eq:td-loss-matrix-form2}) by the following ordinary differental equation (ODE) system:
 \begin{align}\label{eq:td-obj-ode}
    \begin{cases}
        &T_{\phi,z,t} \in \argmin_{T_z} \cL_{\mathrm{td}}(\phi_t, T_z, \psi_t)
        \\ & T_{\psi,z,t} \in \argmin_{T_z} \cL_{\mathrm{td}}(\psi_t, T_z, \phi_t)
        \\ & \dot{\phi}_t = -\nabla_{\phi_t} \cL_{\mathrm{td}}(\phi_t, T_{\phi,z,t}, \psi_t)
        \\ & \dot{\psi}_t = -\nabla_{\psi_t} \cL_{\mathrm{td}}(\psi_t, T_{\psi,z,t}, \phi_t)
    \end{cases}
 \end{align}
 This implicitly assumes that predictors are optimized at a much faster rate than representations -- an important property used in Theorem 1 of \cite{tang2023understanding} to show constant covariance and, thus, no collapse. We now prove this by following similar steps as in the proof of \cite{tang2023understanding}, adapted to our setting. In particular, we prove it for the representations $\phi$ only. Given the symmetry of the losses, the same result can trivially be proven for $\psi$ as well.
 
 We need to prove that $\frac{\di}{\di t}(\phi_t^\transp \phi_t) = 0$. Since $\frac{\di}{\di t}(\phi_t^\transp \phi_t) = (\phi_t^\transp \dot{\phi}_t)^\transp + \phi_t^\transp \dot{\phi}_t$, it is enough to show that $\phi_t^\transp \dot{\phi}_t = 0$. Simple algebra yields
 \begin{align*}
    \nabla_{\phi} \cL_{\mathrm{td}}(\phi, T_z, \psi) = \bE_z \left[ D_\rho \left( \phi T_{z} - U_z \right) T_z^\transp \right],
\end{align*}
\begin{align*}
    \nabla_{T_z} \cL_{\mathrm{td}}(\phi, T_z, \psi) = p(z)\phi^\transp D_\rho \left( \phi T_{z} - U_z \right).
\end{align*}
Therefore,
\begin{align*}
    \phi_t^\transp \dot{\phi}_t &= -\phi_t^\transp \nabla_{\phi_t} \cL_{\mathrm{td}}(\phi_t, T_{\phi,z,t}, \psi_t)
    \\ &= -\phi_t^\transp \bE_z \left[ D_\rho \left( \phi_t T_{\phi,z,t} - U_z \right) T_{\phi,z,t}^\transp \right]
    \\ &= -\sum_{z\in\cZ} p(z) \phi_t^\transp D_\rho \left( \phi_t T_{\phi,z,t} - U_z \right) T_{\phi,z,t}^\transp
    \\ &= -\sum_{z\in\cZ} \nabla_{T_{\phi,z,t}} \cL_{\mathrm{td}}(\phi_t, T_{\phi,z,t}, \psi_t) T_{\phi,z,t}^\transp
    \\ &= 0,
\end{align*}
where the last equation holds since the gradient w.r.t. the predictor is zero at every step (first order optimality conditions from Eq.~\ref{eq:td-obj-ode}).

\subsection{Proof of Theorem \ref{th:td-sp-vs-den-new}}

We begin by rewriting the TD-JEPA loss of Eq.~\ref{eq:asymm.sptd.phi} with the notation of Sec.~\ref{sec:theory} as
\begin{align}\label{eq:td-loss-matrix-form}
    \mathcal{L}_{\text{\algname}}(\phi, T_z, \psi) := \frac{1}{2}\bE_{z\sim\cZ,s\sim\rho,s^+\sim P^{\pi_z}(\cdot | s)} \left[ \Big\| T_{\phi,z}^\transp \phi(s) - \sg{\psi(s^+)} - \gamma  \sg{T_{\phi,z}^\transp \phi(s^+)} \Big\|_2^2 \right].
\end{align} 
Following the proof of Theorem \ref{th:mc-sp-vs-recon}, we can put this in matrix form as
\begin{align}\label{eq:sp-td-matrix}
    \cL_{\mathrm{td}}(\phi, T_z, \psi) := \frac{1}{2}\bE_{z\sim\cZ} \left[ \Big\| D_\rho^{1/2} \left( \phi T_{z} - \sg{P^{\pi_z}\psi} - \gamma  \sg{P^{\pi_z} \phi T_{z}} \right) \Big\|_F^2 \right].
\end{align}   
As we only brought expectations inside the norm, Eq.~\ref{eq:sp-td-matrix} has the same gradients as Eq.~\ref{eq:td-loss-matrix-form}, so we can focus on it to prove the results. Moreover, we can set $D_\rho = I$ by Assumption A2.

\paragraph{Statement 1}

We start by computing the gradients of $\cL_{\mathrm{td}}$ and $\cL_{\mathrm{fw}}$ w.r.t. $T_z$. Up to a multiplicative constant $p(z)$ (which doesn't change the results), we have
\begin{align*}
    \nabla_{T_z} \cL_{\mathrm{td}}(\phi, T_z, \psi) = \phi^\transp \left( \phi T_{z} - {P^{\pi_z}\psi} - \gamma  {P^{\pi_z} \phi T_{z}} \right) = T_z - \phi^\transp P^{\pi_z}\psi - \gamma \phi^\transp P^{\pi_z} \phi T_{z},
\end{align*}
where we used Assumption A1 to set $\phi^\transp \phi = I$. Further using that $\psi^\transp \psi = I$,
\begin{align*}
    \nabla_{T_z} \cL_{\mathrm{fw}}(\phi, T_z, \psi) &= \phi^\transp (\phi T_z \psi^\transp - P^{\pi_z} - \gamma P^{\pi_z} \phi T_z \psi^\transp) \psi
    \\ &= T_z - \phi^\transp P^{\pi_z}\psi - \gamma \phi^\transp P^{\pi_z} \phi T_{z}
    \\ &= \nabla_{T_z} \cL_{\mathrm{td}}(\phi, T_z, \psi).
\end{align*}
Therefore, the gradients w.r.t. $T_z$ of the two losses match, which means that the stationary points (i.e., optimal predictors) are also the same. Setting these gradients to zero we thus find that
\begin{align*}
    T_{\phi,z}^\star = T_{z, \mathrm{fw}}^\star = (\phi^\transp (I - \gamma P^{\pi_z}) \phi)^{-1} \phi^\transp P^{\pi_z}\psi.
\end{align*}
Note that matrix $\phi^\transp (I - \gamma P^{\pi_z}) \phi$ is positive definite and, thus, invertible. This is because $I - \gamma P^{\pi_z}$ is positive definite and $\phi^\transp \phi = I$. Using that $M^{\pi_z} = (I - \gamma P^{\pi_z})^{-1} P^{\pi_z}$,
\begin{align*}
    \phi T_{\phi,z}^\star = \phi T_{z, \mathrm{fw}}^\star = \underbrace{\phi (\phi^\transp (I - \gamma P^{\pi_z}) \phi)^{-1} \phi^\transp (I - \gamma P^{\pi_z})}_{\tilde\Pi_{\phi, z}} M^{\pi_z}\psi,
\end{align*}
where it is easy to verify that $\tilde\Pi_{\phi,z}$ is idempotent ($\tilde\Pi_{\phi,z}\tilde\Pi_{\phi,z} = \tilde\Pi_{\phi,z}$) but not necessarily symmetric, hence an oblique projection as stated.

For the other result in Statement 1, we proceed analogously by first showing that
\begin{align*}
    \nabla_{T_z} \cL_{\mathrm{td}}(\psi, T_z, \phi) = T_z - \psi^\transp P^{\pi_z}\phi - \gamma \psi^\transp P^{\pi_z} \psi T_{z},
\end{align*}
\begin{align*}
    \nabla_{T_z} \cL_{\mathrm{bw}}(\phi, T_z, \psi) = T_z - \psi^\transp (P^{\pi_z})^\transp \phi - \gamma \psi^\transp (P^{\pi_z})^\transp \psi T_{z} = \nabla_{T_z} \cL_{\mathrm{td}}(\psi, T_z, \phi),
\end{align*}
where the last equality is true since $P^{\pi_z}$ is symmetric. Then the result follows as before after equating the gradients to zero, solving for $T_{\psi,z}^\star$, and expressing $\psi T_{\psi,z}^\star$ as a function of $\tilde\Pi_{\psi,z}$.

\paragraph{Statement 2}

We show that the gradients w.r.t. $\phi$ and $\psi$ match for any predictor $T_z$
\begin{align*}
    \nabla_{\phi} \cL_{\mathrm{td}}(\phi, T_z, \psi) = \left( \phi T_{z} - {P^{\pi_z}\psi} - \gamma  {P^{\pi_z} \phi T_{z}} \right) T_z^\transp,
\end{align*}
\begin{align*}
    \nabla_{\phi} \cL_{\mathrm{fw}}(\phi, T_z, \psi) = \left( \phi T_{z}\psi^\transp - {P^{\pi_z}} - \gamma  {P^{\pi_z} \phi T_{z}\psi^\transp} \right) \psi T_z^\transp = \nabla_{\phi} \cL_{\mathrm{td}}(\phi, T_z, \psi),
\end{align*}
where we used that $\psi^\transp \psi = I$. Similarly,
\begin{align*}
    \nabla_{\psi} \cL_{\mathrm{td}}(\psi, T_z, \phi) = \left( \psi T_{z} - {P^{\pi_z}\phi} - \gamma  {P^{\pi_z} \psi T_{z}} \right) T_z^\transp,
\end{align*}
\begin{align*}
    \nabla_{\psi} \cL_{\mathrm{bw}}(\phi, T_z, \psi) = \left( \psi T_{z}\phi^\transp - (P^{\pi_z})^\transp - \gamma  (P^{\pi_z})^\transp \psi T_{z}\phi^\transp \right) \phi T_z^\transp = \nabla_{\psi} \cL_{\mathrm{td}}(\psi, T_z, \phi),
\end{align*}
where we used that $\phi^\transp \phi = I$ and $(P^{\pi_z})^\transp = P^{\pi_z}$. This proves the statement.

\qedwhite

\subsection{Proof of Theorem \ref{th:policy-eval-bellman}}

Let us start from the first inequality. Defining $V_{r}^{\pi_z} \in \bR^{S}$ as a vector containing $V_{r}^{\pi_z}(s)$ for all states $s\in\cS$, we can write the left-hand side for any $r \in \bR^{S}$ as
\begin{align*}
    \bE_{z\in\cZ} \left[ \sum_{s\in\cS}\left( V_{r}^{\pi_z}(s) - \phi(s)^\transp T_z \omega_r \right)^2 \right] = \bE_{z\in\cZ} \|  V_{r}^{\pi_z} - \phi T_z \omega_r  \|_2^2.
\end{align*}
Since $V_{r}^{\pi_z} = M^{\pi_z}r$ and, by Assumption A1, $\omega_r = \psi^\transp r$,
\begin{align*}
    \bE_{z\in\cZ} \| V_{r}^{\pi_z} - \phi T_z \omega_r \|_2^2 &= \bE_{z\in\cZ} \| M^{\pi_z}r - \phi T_z \psi^\transp r \|_2^2
    \\ &\leq \bE_{z\in\cZ} \| M^{\pi_z} - \phi T_z \psi^\transp \|_F^2 \| r \|_2^2
    \\ &= 2 \cL_{\mathrm{SM}}(\phi, T_z, \psi) \| r \|_2^2.
\end{align*}
The inequality is thus obtained by maximizing both sides over rewards with norm bounded by $1$.

Let us now prove the bounds of $\cL_{\mathrm{SM}}$ in terms of the Bellman errors $\cL_{\mathrm{fw}}$ and $\cL_{\mathrm{bw}}$. Let us fix $z\in\cZ$ and recall that $M^{\pi_z}$ admits both a ``forward'' Bellman equation, $M^{\pi_z} = P^{\pi_z} + \gamma P^{\pi_z}M^{\pi_z}$, and a ``backward'' one, $M^{\pi_z} = P^{\pi_z} + \gamma M^{\pi_z} P^{\pi_z}$ \citep{blier2021learning}. This implies that $M^{\pi_z} = (I - \gamma P^{\pi_z})^{-1} P^{\pi_z} = P^{\pi_z} (I - \gamma P^{\pi_z})^{-1} $. Then, for any matrix $M \in \bR^{S\times S}$,
    \begin{align*}
        M - M^{\pi_z} &= (I-\gamma P^{\pi_z})^{-1} \left( (I-\gamma P^{\pi_z}) M - P^{\pi_z} \right) = (I-\gamma P^{\pi_z})^{-1} \left( M - P^{\pi_z} - \gamma P^{\pi_z} M \right),
        \\ M - M^{\pi_z} &= \left( M (I-\gamma P^{\pi_z}) - P^{\pi_z} \right) (I-\gamma P^{\pi_z})^{-1} = \left( M - P^{\pi_z} - \gamma M P^{\pi_z} \right) (I-\gamma P^{\pi_z})^{-1}.
    \end{align*}
    Using the first set of equalities with $M = \phi T_z \psi^\transp$, we can easily bound
    \begin{align*}
        \cL_{\mathrm{SM}}(\phi, T_z, \psi) &= \frac{1}{2}\bE_{z \sim \cZ}\left[ \| \phi T_z \psi^\transp - M^{\pi_z} \|_F^2 \right]
        \\ &\leq \frac{1}{2}\bE_{z \sim \cZ}\left[ \| (I-\gamma P^{\pi_z})^{-1} \|_2^2 \| \phi T_z \psi^\transp - P^{\pi_z} - \gamma P^{\pi_z} \phi T_z \psi^\transp \|_F^2 \right],
    \end{align*}
    where we used the inequality $\|XY\|_F^2 \leq \|X\|_2^2 \|Y\|_F^2$ with $\|\cdot\|_2^2$ denoting the operator norm and $\|\cdot\|_F^2$ the frobenius norm. Moreover,
    \begin{align*}
        \| (I-\gamma P^{\pi_z})^{-1} \|_2^2 = \frac{1}{(1-\gamma)^2} \| (1-\gamma)(I-\gamma P^{\pi_z})^{-1} \|_2^2 \leq \frac{S}{(1-\gamma)^2},
    \end{align*}
    where the last inequality holds since $(1-\gamma)(I-\gamma P^{\pi_z})^{-1}$ is a stochastic matrix. Hence,
    \begin{align*}
        \cL_{\mathrm{SM}}(\phi, T_z, \psi) \leq \frac{S}{(1-\gamma)^2} \cL_{\mathrm{fw}}(\phi, T_z, \psi),
    \end{align*}
    which proves the first inequality with $c = \frac{S}{(1-\gamma)^2}$. The second one can be proved analogously.

\qedwhite

\section{Theoretical Analysis under Relaxed Assumptions}\label{app:theory-general}

This section describes how the main assumptions used in Section \ref{sec:theory} can be removed, namely the uniform state distribution (A1), identity covariances (A2), and symmetric kernels $P^{\pi_z}$ or $M^{\pi_z}$ (A3). We shall derive similar results Th.~\ref{th:mc-sp-vs-recon} and Th.~\ref{th:td-sp-vs-den-new}, but at the price of more complex proofs and notation. We do so through a novel ``gradient matching'' argument that reduces a general latent-predictive loss to density approximation. As we shall see, this encompasses not only MC-JEPA and TD-JEPA, but also existing methods \citep{tang2023understanding,khetarpal2024unifying,lawson2025self}.

\subsection{Reduction from latent-predictive to density-based losses}

Let $\cZ$ be a finite set. For $z\in\cZ$, let $\Xi_z \in \bR^{S \times S}$ be a generic kernel. For instance, $\Xi_z$ may be the successor measure $M^{\pi_z}$ or the one-step kernel $P^{\pi_z}$, but it is not important at this point. Using the same notation as Section \ref{sec:theory} and Appendix \ref{app:proofs}, consider the following density-based loss:
\begin{align}
    \cL_{\mathrm{dens}}(\phi, T_z, \psi) :&= \frac{1}{2}\bE_{z \sim \cZ, s\sim \rho, s^+ \sim \rho}\left[ \left( \phi(s)^\transp T_z \psi(s^+) - \frac{\Xi_z(s^+ | s)}{\rho(s^+)} \right)^2 \right]\notag
    \\ &= \frac{1}{2}\bE_{z \sim \cZ}\left[ \| D_{\rho}^{1/2} (\phi T_z \psi^\transp - \Xi_z D_{\rho}^{-1}) D_{\rho}^{1/2} \|_F^2 \right]. \label{eq:density-loss}
\end{align}
Minimizing this loss over representations $\phi,\psi$ and a collection of predictors $\{T_z\}_{z\in\cZ}$ is equivalent to finding the best multilinear approximation to the densities of the $\Xi_z$ w.r.t.\ the state distribution $\rho$. Note that this is a well-defined loss (i.e., it does not involve stop-gradient operations) and the prediction targets $\Xi_z D_\rho^{-1}$ are not a function of the representations being learned. Our goal is to show that certain latent-predictive dynamics optimize this density-based loss.

For didactic purpose, let us consider two abstract latent-predictive losses $\cL_{\phi}(\phi, T_{\phi,z}, \psi)$ and $\cL_{\psi}(\psi, T_{\psi,z}, \phi)$. We shall specify what these are later. $\cL_{\phi}(\phi, T_{\phi,z}, \psi)$ is optimized over $\phi$ and $T_{\phi,z}$ for all $z$, while $\cL_{\psi}(\psi, T_{\psi,z}, \phi)$ is optimized over $\psi$ and $T_{\psi,z}$ for all $z$. As common in the literature, we study a continuous-time relaxation of gradient descend dynamics by assuming that predictors are optimized at a much faster rate than representations -- an important property to ensure that the latter ones do not collapse \citep{tang2023understanding}. This process can be described by the following ordinary differental equation (ODE) system:
\begin{align}\label{eq:general-obj-ode}
    \begin{cases}
        &T_{\phi,z,t} \in \argmin_{T_z} \cL_{\phi}(\phi_t, T_z, \psi_t)
        \\ & T_{\psi,z,t} \in \argmin_{T_z} \cL_{\psi}(\psi_t, T_z, \phi_t)
        \\ & \dot{\phi}_t = -\nabla_{\phi_t} \cL_{\phi}(\phi_t, T_{\phi,z,t}, \psi_t)
        \\ & \dot{\psi}_t = -\nabla_{\psi_t} \cL_{\psi}(\psi_t, T_{\psi,z,t}, \phi_t)
    \end{cases}
\end{align}
We now ask the question: how should $\cL_{\phi}$ and $\cL_{\psi}$ be defined for the dynamics of Eq.~\ref{eq:general-obj-ode} to optimize the density-based loss of Eq.~\ref{eq:density-loss}? The following result provides and answer.
\begin{metaframe}
\begin{theorem}\label{th:lyapunov}
    Suppose that the gradients of $\cL_{\mathrm{dens}}$, $\cL_{\phi}$, and $\cL_{\psi}$ match for all $\phi$, $\psi$, $T_z$, i.e.,
    \begin{enumerate}
        \item $\nabla_{T_z} \cL_{\mathrm{dens}}(\phi, T_z, \psi) = \nabla_{T_z} \cL_{\phi}(\phi, T_z, \psi)$ 
        \item $\nabla_{T_z} \cL_{\mathrm{dens}}(\phi, T_z^\transp, \psi) = \nabla_{T_z} \cL_{\psi}(\psi, T_z, \phi)$ 
        \item $\nabla_{\phi} \cL_{\mathrm{dens}}(\phi, T_z, \psi) = \nabla_{\phi} \cL_{\phi}(\phi, T_z, \psi)$ 
        \item $\nabla_{\psi} \cL_{\mathrm{dens}}(\phi, T_z^\transp, \psi) = \nabla_{\psi} \cL_{\psi}(\psi, T_z, \phi)$ 
    \end{enumerate}
    Then, $\cL_{\mathrm{dens}}$ is a Lyapunov function for the ODE of Eq.~\ref{eq:general-obj-ode}.
\end{theorem}
\end{metaframe}
\begin{proof}
    Let $T_{z,t} \in \argmin_{T_z} \cL_{\mathrm{dens}}(\phi_t, T_z, \psi_t)$ be the optimal predictor for the density-based loss given $(\phi_t, \psi_t)$ and define $\cL(t) := \cL_{\mathrm{dens}}(\phi_t, T_{z,t}, \psi_t)$. We verify that a $\cL(t)$, and thus $\cL_{\mathrm{dens}}$, is a Lyapunov function for the ODE of Eq.~\ref{eq:general-obj-ode}. First note that $\cL_{\mathrm{dens}}$ is continuous and has continuous first derivates. By the chain rule, 
    \begin{align*}
        \frac{\di}{\di t} \cL(t) 
        &= \nabla_\phi \cL_{\mathrm{dens}}(\phi_t, T_{z,t}, \psi_t) \cdot \dot \phi_t + \nabla_\psi \cL_{\mathrm{dens}}(\phi_t, T_{z,t}, \psi_t) \cdot \dot \psi_t,
    \end{align*}
    where the $\cdot$ operation is the dot product of the vectorized matrices.
    By Eq.~\ref{eq:general-obj-ode}, $\dot \phi_t = -\nabla_{\phi} \cL_{\phi}(\phi_t, T_{\phi,z,t}, \psi_t)$ and $\dot \psi_t = -\nabla_{\psi} \cL_{\psi}(\psi_t, T_{\psi,z,t}, \phi_t) $. Moreover, assumptions 1 and 2 directly yield $T_{\phi,z,t} = T_{z,t}$ and $T_{\psi,z,t} = T_{z,t}^\transp$. Plugging these into $\dot \phi_t$ and $\dot \psi_t$ and using assumptions 3 and 4,
    \begin{align*}
        \dot \phi_t &= - \nabla_{\phi} \cL_{\phi}(\phi_t, T_{z,t}, \psi_t) = - \nabla_\phi \cL_{\mathrm{dens}}(\phi_t, T_{z,t}, \psi_t),
        \\ \dot \psi_t &= -\nabla_{\psi} \cL_{\psi}(\psi_t, T_{z,t}^\transp, \phi_t) = - \nabla_\psi \cL_{\mathrm{dens}}(\phi_t, T_{z,t}, \psi_t).
    \end{align*}
    Thus,
    \begin{align*}
        \frac{\di}{\di t} \cL(t) 
        = - \| \nabla_\phi \cL_{\mathrm{dens}}(\phi_t, T_{z,t}, \psi_t) \|_F^2 - \| \nabla_\psi \cL_{\mathrm{dens}}(\phi_t, T_{z,t}, \psi_t) \|_F^2 \leq 0,
    \end{align*}
     Moreover, we clearly have $\frac{\di}{\di t} \cL(t) < 0$ if $(\phi_t,\psi_t)$ is not a stationary point of the ODE, which proves the statement.
\end{proof}

Therefore, if the gradients of $\cL_{\mathrm{dens}}$, $\cL_{\phi}$, and $\cL_{\psi}$ match, the gradient dynamics of Eq.~\ref{eq:general-obj-ode} monotonically improve the density-based loss over time. We remark that this does not imply convergence to the global minimum as $\cL_{\mathrm{dens}}$ is not convex. Th.~\ref{th:lyapunov} thus suggests a simple trick for designing the right latent-predictive dynamics to optimize a given density-based loss: just find $\cL_{\phi}$ and $\cL_{\psi}$ whose gradients match those of $\cL_{\mathrm{dens}}$. Latent predictive losses that satisfy this property are derived in the next result.
\begin{metaframe}
\begin{proposition}\label{prop:gradient-matching}
    Consider the following latent-predictive losses
    \begin{align}
        \cL_{\phi}(\phi, T_z, \psi) &:= \frac{1}{2}\bE_{z\sim\cZ} \| D_\rho^{1/2} (\phi T_z - \sg{\Xi_z \psi \Sigma_\psi^{-1}}) \|_{\sg{\Sigma}_\psi}^2, \label{eq:general-fw-loss}
        \\ \cL_{\psi}(\psi, T_z, \phi) &:= \frac{1}{2}\bE_{z\sim\cZ} \| D_{\rho}^{1/2} (\psi T_z - \sg{\Xi_z^* \phi \Sigma_\phi^{-1}}) \|_{\sg{\Sigma}_\phi}^2, \label{eq:general-bw-loss}
    \end{align}
    where $\| X \|_W = \| X W^{1/2} \|_F$, $\Sigma_\phi := \phi^\transp D_{\rho} \phi$, $\Sigma_{\psi} := \psi^\transp D_{\rho} \psi$, and $\Xi_z^* := D_{\rho}^{-1} \Xi_z^\transp D_{\rho}$ is the $\rho$-adjoint of $\Xi_z$. Then, $\cL_{\phi}$ and $\cL_{\psi}$ satisfy the gradient matching conditions for $\cL_{\mathrm{dens}}$ stated in Th.~\ref{th:lyapunov}.
\end{proposition}
\end{metaframe}
\begin{proof}
    This can be directly obtained through simple linear algebra.
\end{proof}

Note that, while the latent-predictive losses of Eq. \ref{eq:general-fw-loss} and \ref{eq:general-bw-loss} are expressed in expectation w.r.t. the kernels $\Xi_z$ and $\Xi_z^*$, sample-based estimators for both are possible. In particular, while Eq. \ref{eq:general-fw-loss} involves sampling from the kernel $\Xi_z$, Eq.~\ref{eq:general-bw-loss} involves a ``backward'' sampling operation given by the adjoint of $\Xi_z$.

\subsection{Generalizing existing results}\label{app:generalize-existing}

Through the right assumptions and choice of $\Xi_z$, Proposition \ref{prop:gradient-matching} and Theorem \ref{th:lyapunov} yield several existing results. First, with $\Xi_z = M^{\pi_z}$, we note that Eq. \ref{eq:general-fw-loss} and \ref{eq:general-bw-loss} are equivalent to MC-JEPA (cf. Eq.~\ref{eq:mc-fw-sp-loss}) with additional covariance weighting and a backward-in-time sampling in the loss for $\psi$. Assuming that $D_\rho = I$ (A1), $\Sigma_\phi = \Sigma_\psi = I$ (A2), and $P^{\pi_z}$ is symmetric for all $z$ (A3), this mismatch is resolved and we recover Th.~\ref{th:mc-sp-vs-recon} under its same assumptions. On the other hand, Proposition \ref{prop:gradient-matching} and Theorem \ref{th:lyapunov} show what happens when such assumptions are relaxed: if only A3 holds\footnote{To be precise, here we need the density $M^{\pi_z}D_\rho^{-1}$ to be symmetric, which implies that $(M^{\pi_z})^* = M^{\pi_z}$.}, then Th.~\ref{th:mc-sp-vs-recon} still holds up to covariance transformations, while if A3 is further removed then MC-JEPA needs to be modified with backward sampling in the loss of $\psi$ to guarantee optimization of the successor measure approximation loss.

Several analyses in related works, which all assume A1-A3, are also recovered by proper choice of $\Xi_z$. For instance, for a single policy $\pi$ (i.e., $|\cZ|=1$), we obtain Theorem 6 of \cite{tang2023understanding} by setting $\Xi_z = P^\pi$ and Theorem 4.1 of \cite{lawson2025self} by setting $\Xi_z = M^{\pi}$. By letting $\cZ$ correspond to the action space $\cA$, we recover Theorem 2 of \cite{khetarpal2024unifying} by setting $\Xi_a = P^{\pi}_a$ for all actions $a$, with $P_a$ the matrix containing $P(s'|s,a)$ for all $(s,s')$.

\subsection{TD-JEPA with forward-backward-in-time sampling}
\label{app:forward-backward}

In the previous section, we have seen that MC-JEPA requires a backward-in-time sampling operation in the loss for $\psi$ to provably optimize the successor measure approximation loss. In this section, we derive the TD dynamics corresponding to this process. 

We thus start from $\cL_{\mathrm{dens}}$, $\cL_{\phi}$, and $\cL_{\psi}$ with $\Xi_z = M^{\pi_z}$ and replace the latter with the one-step kernel $P^{\pi_z}$ plus the bootstrapped parameterization $M^{\pi_z} \simeq \phi T_z \psi^\transp D_{\rho}$. For $\cL_{\mathrm{dens}}$, similarly to Th.~\ref{th:td-sp-vs-den-new}, we derive density-based forward and backward TD losses, respectively based on the application of the forward and backward Bellman operator. For the first one, we use that $M^{\pi_z} = P^{\pi_z} + \gamma P^{\pi_z} M^{\pi_z} \simeq P^{\pi_z} + \gamma P^{\pi_z} \phi T_z \psi^\transp D_{\rho}$ and write
\begin{align}\label{eq:general-td-dens-fw}
    \cL_{\mathrm{fw}}(\phi, T_z, \psi) &:= \frac{1}{2}\bE_{z \sim \cZ}\left[ \| D_\rho^{1/2}(\phi T_z \psi^\transp - P^{\pi_z}D_\rho^{-1} - \gamma \sg{P^{\pi_z} \phi T_z \psi^\transp}) D_\rho^{1/2} \|_F^2 \right].
\end{align}
For the second one, we rewrite the backward Bellman equation in terms of the adjoint of $M^{\pi_z}$ as
\begin{align*}
    (M^{\pi_z})^* &= (P^{\pi_z})^* + \gamma (M^{\pi_z} P^{\pi_z})^*
    \\ &= (P^{\pi_z})^* + \gamma D_{\rho}^{-1} (P^{\pi_z})^\transp (M^{\pi_z})^\transp D_\rho
    \\ &\simeq (P^{\pi_z})^* + \gamma D_{\rho}^{-1} (P^{\pi_z})^\transp D_{\rho} \psi T_z^\transp \phi^\transp D_\rho
    \\ &= (P^{\pi_z})^* + \gamma (P^{\pi_z})^* \psi T_z^\transp \phi^\transp D_\rho.
\end{align*}
This yields
\begin{align}\label{eq:general-td-dens-bw}
    \cL_{\mathrm{bw}}(\phi, T_z, \psi) &:= \frac{1}{2}\bE_{z \sim \cZ}\left[ \| D_\rho^{1/2}(\psi T_z \phi^\transp - (P^{\pi_z})^* D_\rho^{-1}  - \gamma (P^{\pi_z})^* \sg{\psi T_z \phi^\transp}) D_\rho^{1/2} \|_F^2 \right].
\end{align}
We then use the same bootstrapping for the latent-predictive losses of Eq.~\ref{eq:general-fw-loss} and \ref{eq:general-bw-loss}, thus obtaining
    \begin{align}
        \cL_{\phi}(\phi, T_z, \psi) &:= \frac{1}{2}\bE_{z\sim\cZ} \| D_\rho^{1/2} (\phi T_z - \sg{P^{\pi_z} \psi \Sigma_\psi^{-1}} - \gamma \sg{P^{\pi_z} \phi T_z}) \|_{\sg{\Sigma}_\psi}^2, \label{eq:general-td-fw-loss}
        \\ \cL_{\psi}(\psi, T_z, \phi) &:= \frac{1}{2}\bE_{z\sim\cZ} \| D_{\rho}^{1/2} (\psi T_z - \sg{(P^{\pi_z})^* \phi \Sigma_\phi^{-1}} - \gamma \sg{(P^{\pi_z})^* \psi T_z}) \|_{\sg{\Sigma}_\phi}^2. \label{eq:general-td-bw-loss}
    \end{align}
As for MC-JEPA, this is the counterpart of TD-JEPA with additional covariance weighting and backward-in-time sampling in the loss of $\psi$. Through simple algebra, it is not difficult to check that $\cL_{\phi}$ and $\cL_{\psi}$ have matching gradients (as in Th.~\ref{th:lyapunov}) w.r.t. $\cL_{\mathrm{fw}}$ and $\cL_{\mathrm{bw}}$, respectively. This implies that Th.~\ref{th:td-sp-vs-den-new} holds for this modified algorithm without assumptions A1-A3.

Unfortunately, differently from the original TD-JEPA, this variant is not easy to be optimized off-policy. In fact, while for Eq.~\ref{eq:general-td-fw-loss} we can simply replace the on-policy kernel $P^{\pi_z}$ with $P(s'|s,a)$ and condition the predictor on actions, the same trick cannot be used for the adjoint $(P^{\pi_z})^*$ in Eq.~\ref{eq:general-td-bw-loss}. This is because there is no action-conditioned backward Bellman equation. We leave the study of practical learning dynamics for this theoretically sound variant of TD-JEPA for future work.

\section{Additional Results}
\label{app:exp}

This section reports additional experiments, as well as detailed numerical results for plots in the main part of the paper.

\subsection{Explicit state encoders for zero-shot baselines}
\label{app:exp.state_encoder}

As mentioned in Section~\ref{sec:exp}, existing zero-shot methods do not necessarily learn explicit state embeddings \cite{wu2018laplacian, touati2021learning, park2024foundation}.
Nevertheless, we find that introducing a state encoder tends to improve average zero-shot performance, even in the absence of a specific representation learning objective.
We consider three established zero-shot baselines (Laplacian \citep{wu2018laplacian}, FB \citep{touati2021learning} and HILP \citep{Park25ogbench}), and compare their performance without an explicit state encoder (i.e., successor features are trained on raw states), to those attained when introducing a state encoder (either shallow or deep). Note that, in this case, the state encoder is trained through gradient flowing from the original loss, and is not coupled to an ad-hoc objective. Results are summarized in Tab.~\ref{tab:state_encoder}, which only considers proprioceptive domains as a state encoder is necessary when learning from pixels.
We observe that, while having an explicit state encoder may be detrimental in few specific domains, it remains beneficial on average, and crucial to obtain better performance in some domains.
The optimal depth of the encoder is however domain-specific: OGBench domains generally prefer a deeper encoder, while DMC domains can be solved with a shallow encoder, or no explicit encoder at all. In order to maximize performance of the baselines, we thus evaluate them coupled with a deep encoder in OGBench, and with a shallow one in DMC (e.g., in Tab.~\ref{tab:main.results}).
We additionally summarize performance differences induced by this choice of encoders in Fig.~\ref{fig:state_encoder} (\emph{left}).

\begin{table}
\centering
\resizebox{\textwidth}{!}{
\begin{tabular}{l@{\hspace{30pt}}ccc@{\hspace{30pt}}ccc@{\hspace{30pt}}ccc}
\toprule
 & FB & FB$_{\text{shallow}}$ & FB$_{\text{deep}}$ & HILP & HILP$_{\text{shallow}}$ & HILP$_{\text{deep}}$ & Laplacian & Laplacian$_{\text{shallow}}$ & Laplacian$_{\text{deep}}$ \\ \midrule
DMC (\texttt{avg}) & \textbf{648.9 \scriptsize{$\pm$ 7.5}} & \textbf{648.2 \scriptsize{$\pm$ 4.1}} & 632.0 \scriptsize{$\pm$ 4.4} & \textbf{659.4 \scriptsize{$\pm$ 4.1}} & 620.1 \scriptsize{$\pm$ 8.4} & 603.2 \scriptsize{$\pm$ 5.8} & \textbf{585.6 \scriptsize{$\pm$ 8.9}} & \textbf{591.1 \scriptsize{$\pm$ 10.7}} & \textbf{598.8 \scriptsize{$\pm$ 7.5}} \\
\quad \texttt{walker} & \scriptsize{788.7} \scriptsize{$\pm$ 4.3} & \textbf{\scriptsize{811.5} \scriptsize{$\pm$ 5.9}} & \textbf{\scriptsize{812.4} \scriptsize{$\pm$ 12.9}} & \textbf{\scriptsize{783.6} \scriptsize{$\pm$ 8.4}} & \textbf{\scriptsize{796.4} \scriptsize{$\pm$ 7.7}} & \textbf{\scriptsize{785.0} \scriptsize{$\pm$ 9.1}} & \textbf{\scriptsize{754.6} \scriptsize{$\pm$ 16.7}} & \textbf{\scriptsize{769.7} \scriptsize{$\pm$ 4.7}} & \textbf{\scriptsize{762.1} \scriptsize{$\pm$ 7.8}} \\
\quad \texttt{cheetah} & \textbf{\scriptsize{662.8} \scriptsize{$\pm$ 5.7}} & \textbf{\scriptsize{672.7} \scriptsize{$\pm$ 4.9}} & \scriptsize{635.7} \scriptsize{$\pm$ 22.3} & \textbf{\scriptsize{635.2} \scriptsize{$\pm$ 9.4}} & \scriptsize{618.3} \scriptsize{$\pm$ 5.8} & \scriptsize{612.3} \scriptsize{$\pm$ 12.2} & \textbf{\scriptsize{640.5} \scriptsize{$\pm$ 9.1}} & \textbf{\scriptsize{614.5} \scriptsize{$\pm$ 18.9}} & \textbf{\scriptsize{630.7} \scriptsize{$\pm$ 17.9}} \\
\quad \texttt{quadruped} & \scriptsize{574.1} \scriptsize{$\pm$ 11.3} & \textbf{\scriptsize{595.6} \scriptsize{$\pm$ 9.1}} & \textbf{\scriptsize{590.6} \scriptsize{$\pm$ 10.6}} & \textbf{\scriptsize{695.8} \scriptsize{$\pm$ 13.5}} & \textbf{\scriptsize{694.8} \scriptsize{$\pm$ 11.0}} & \textbf{\scriptsize{681.1} \scriptsize{$\pm$ 16.2}} & \scriptsize{590.6} \scriptsize{$\pm$ 31.0} & \textbf{\scriptsize{635.0} \scriptsize{$\pm$ 38.7}} & \textbf{\scriptsize{651.5} \scriptsize{$\pm$ 12.7}} \\
\quad \texttt{pointmass} & \textbf{\scriptsize{570.0} \scriptsize{$\pm$ 22.6}} & \scriptsize{513.0} \scriptsize{$\pm$ 20.0} & \scriptsize{489.3} \scriptsize{$\pm$ 14.8} & \textbf{\scriptsize{522.8} \scriptsize{$\pm$ 18.9}} & \scriptsize{371.0} \scriptsize{$\pm$ 37.1} & \scriptsize{334.5} \scriptsize{$\pm$ 27.9} & \textbf{\scriptsize{356.7} \scriptsize{$\pm$ 24.0}} & \textbf{\scriptsize{345.1} \scriptsize{$\pm$ 22.4}} & \textbf{\scriptsize{351.0} \scriptsize{$\pm$ 28.5}} \\
\midrule
OGBench (\texttt{avg}) & 19.07 \scriptsize{$\pm$ 0.65} & 21.96 \scriptsize{$\pm$ 0.81} & \textbf{39.04 \scriptsize{$\pm$ 0.66}} & 30.51 \scriptsize{$\pm$ 1.20} & 35.02 \scriptsize{$\pm$ 0.92} & \textbf{37.98 \scriptsize{$\pm$ 1.11}} & 9.07 \scriptsize{$\pm$ 0.71} & 9.29 \scriptsize{$\pm$ 0.59} & \textbf{14.81 \scriptsize{$\pm$ 1.32}} \\
\quad \texttt{antmaze-mn} & \scriptsize{45.20} \scriptsize{$\pm$ 2.05} & \scriptsize{49.00} \scriptsize{$\pm$ 2.13} & \textbf{\scriptsize{73.00} \scriptsize{$\pm$ 2.72}} & \scriptsize{62.20} \scriptsize{$\pm$ 2.39} & \scriptsize{60.22} \scriptsize{$\pm$ 2.57} & \textbf{\scriptsize{83.60} \scriptsize{$\pm$ 2.63}} & \scriptsize{17.00} \scriptsize{$\pm$ 2.72} & \scriptsize{25.80} \scriptsize{$\pm$ 3.48} & \textbf{\scriptsize{50.00} \scriptsize{$\pm$ 4.94}} \\
\quad \texttt{antmaze-ln} & \scriptsize{19.80} \scriptsize{$\pm$ 2.62} & \scriptsize{15.60} \scriptsize{$\pm$ 2.54} & \textbf{\scriptsize{36.80} \scriptsize{$\pm$ 4.28}} & \scriptsize{34.00} \scriptsize{$\pm$ 2.86} & \scriptsize{27.40} \scriptsize{$\pm$ 4.00} & \textbf{\scriptsize{52.60} \scriptsize{$\pm$ 3.86}} & \scriptsize{13.60} \scriptsize{$\pm$ 2.68} & \scriptsize{10.60} \scriptsize{$\pm$ 2.46} & \textbf{\scriptsize{21.60} \scriptsize{$\pm$ 3.90}} \\
\quad \texttt{antmaze-ms} & \scriptsize{20.60} \scriptsize{$\pm$ 5.38} & \scriptsize{32.40} \scriptsize{$\pm$ 3.18} & \textbf{\scriptsize{70.40} \scriptsize{$\pm$ 3.95}} & \scriptsize{12.00} \scriptsize{$\pm$ 3.88} & \textbf{\scriptsize{43.60} \scriptsize{$\pm$ 5.12}} & \textbf{\scriptsize{50.60} \scriptsize{$\pm$ 2.46}} & \textbf{\scriptsize{14.20} \scriptsize{$\pm$ 3.24}} & \scriptsize{12.60} \scriptsize{$\pm$ 3.31} & \textbf{\scriptsize{21.40} \scriptsize{$\pm$ 4.32}} \\
\quad \texttt{antmaze-ls} & \scriptsize{8.00} \scriptsize{$\pm$ 2.65} & \scriptsize{15.80} \scriptsize{$\pm$ 1.05} & \textbf{\scriptsize{49.80} \scriptsize{$\pm$ 5.64}} & \scriptsize{4.20} \scriptsize{$\pm$ 2.05} & \textbf{\scriptsize{11.20} \scriptsize{$\pm$ 1.04}} & \textbf{\scriptsize{12.20} \scriptsize{$\pm$ 1.75}} & \scriptsize{4.44} \scriptsize{$\pm$ 2.30} & \scriptsize{4.00} \scriptsize{$\pm$ 1.58} & \textbf{\scriptsize{11.80} \scriptsize{$\pm$ 1.47}} \\
\quad \texttt{antmaze-me} & \scriptsize{23.00} \scriptsize{$\pm$ 3.20} & \scriptsize{25.80} \scriptsize{$\pm$ 1.65} & \textbf{\scriptsize{51.60} \scriptsize{$\pm$ 2.65}} & \textbf{\scriptsize{6.67} \scriptsize{$\pm$ 2.16}} & \textbf{\scriptsize{8.60} \scriptsize{$\pm$ 1.33}} & \scriptsize{2.00} \scriptsize{$\pm$ 0.84} & \textbf{\scriptsize{0.40} \scriptsize{$\pm$ 0.40}} & \textbf{\scriptsize{1.00} \scriptsize{$\pm$ 0.68}} & \textbf{\scriptsize{0.80} \scriptsize{$\pm$ 0.61}} \\
\quad \texttt{cube-single} & \scriptsize{21.00} \scriptsize{$\pm$ 2.52} & \scriptsize{26.80} \scriptsize{$\pm$ 2.62} & \textbf{\scriptsize{49.60} \scriptsize{$\pm$ 3.83}} & \textbf{\scriptsize{84.00} \scriptsize{$\pm$ 2.76}} & \textbf{\scriptsize{88.00} \scriptsize{$\pm$ 2.49}} & \scriptsize{74.20} \scriptsize{$\pm$ 3.53} & \textbf{\scriptsize{19.20} \scriptsize{$\pm$ 2.11}} & \textbf{\scriptsize{18.20} \scriptsize{$\pm$ 2.14}} & \scriptsize{15.11} \scriptsize{$\pm$ 1.49} \\
\quad \texttt{cube-double} & \textbf{\scriptsize{4.20} \scriptsize{$\pm$ 0.96}} & \textbf{\scriptsize{4.00} \scriptsize{$\pm$ 0.79}} & \scriptsize{2.60} \scriptsize{$\pm$ 0.43} & \textbf{\scriptsize{27.11} \scriptsize{$\pm$ 3.02}} & \textbf{\scriptsize{29.20} \scriptsize{$\pm$ 2.83}} & \scriptsize{20.00} \scriptsize{$\pm$ 2.72} & \textbf{\scriptsize{2.80} \scriptsize{$\pm$ 0.53}} & \textbf{\scriptsize{3.40} \scriptsize{$\pm$ 0.85}} & \textbf{\scriptsize{2.00} \scriptsize{$\pm$ 0.42}} \\
\quad \texttt{scene} & \textbf{\scriptsize{23.80} \scriptsize{$\pm$ 2.28}} & \textbf{\scriptsize{22.00} \scriptsize{$\pm$ 2.29}} & \scriptsize{12.80} \scriptsize{$\pm$ 1.61} & \textbf{\scriptsize{42.20} \scriptsize{$\pm$ 1.75}} & \textbf{\scriptsize{45.00} \scriptsize{$\pm$ 2.88}} & \textbf{\scriptsize{43.80} \scriptsize{$\pm$ 1.90}} & \textbf{\scriptsize{7.80} \scriptsize{$\pm$ 1.31}} & \textbf{\scriptsize{5.80} \scriptsize{$\pm$ 0.96}} & \textbf{\scriptsize{7.80} \scriptsize{$\pm$ 1.28}} \\
\quad \texttt{puzzle-3x3} & \textbf{\scriptsize{6.00} \scriptsize{$\pm$ 0.84}} & \textbf{\scriptsize{6.22} \scriptsize{$\pm$ 0.97}} & \textbf{\scriptsize{4.80} \scriptsize{$\pm$ 0.68}} & \textbf{\scriptsize{2.20} \scriptsize{$\pm$ 0.55}} & \textbf{\scriptsize{2.00} \scriptsize{$\pm$ 0.52}} & \textbf{\scriptsize{2.80} \scriptsize{$\pm$ 0.68}} & \textbf{\scriptsize{2.20} \scriptsize{$\pm$ 0.63}} & \textbf{\scriptsize{2.20} \scriptsize{$\pm$ 0.63}} & \textbf{\scriptsize{2.80} \scriptsize{$\pm$ 0.68}} \\
\bottomrule
\end{tabular}
}
\caption{Ablation over encoder depth on proprioceptive DMC and OGBench domains for zero-shot baselines. Each method is evaluated in three variants: a baseline without explicit state encoder, one with a \textit{shallow} (i.e., linear) state  encoder, one with a \textit{deep} one (2 or 4 hidden layers for DMC and OGBench, respectively, see Tab.~\ref{tab:hyperparameters}.) We report mean performance and standard error. Results for each variant are bold if their confidence intervals overlap with the best variant of the same method.}
\label{tab:state_encoder}
\end{table}

\subsection{Contrastive variant of symmetric \algname}

\begin{figure}
    \vspace{-8pt}
    \centering
    \begin{minipage}{.5\textwidth}
    \centering
    \includegraphics[width=\textwidth]{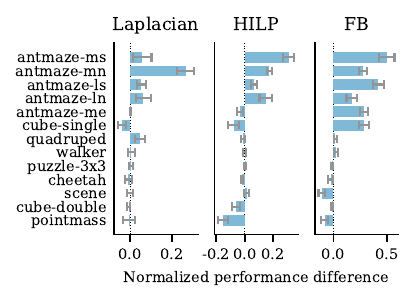}
    \end{minipage}%
    \begin{minipage}{0.5\textwidth}
    \centering
    \includegraphics[width=\textwidth]{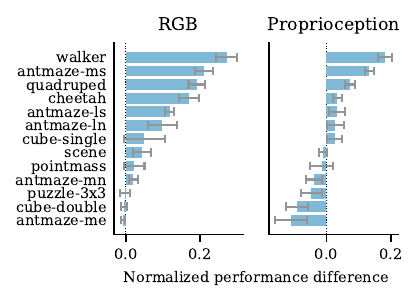}
    \end{minipage}%
    \vspace{-8pt}
    \caption{Difference in normalized performance between zero-shot baselines with and without an explicit encoder (\textbf{left}); normalized performance difference between symmetric \algname and its contrastive variant (\textbf{right}). Error bars represent standard errors on normalized performance differences. \looseness -1}
    \label{fig:state_encoder}
    \vspace{-8pt}
\end{figure}

In order to further isolate the effect of different objectives on zero-shot performance, we instantiate the symmetric variant of \algname described by Eq.~\ref{eq:td-obj}, where a single representation is used both as state and task encoder, as well as its contrastive counterpart, which uses an objective similar to that in \citet{touati2021learning}. These algorithms are described in Alg.~\ref{alg:sptd_sym}, where \textcolor{our_blue}{blue} and \textcolor{our_orange}{yellow} lines are exclusive to the latent-predictive and contrastive variants, respectively.
Intuitively, the contrastive variant can be seen as a specific, symmetric instantiation of FB \citep{touati2021learning}, and the comparison to the symmetric variant parallels that between \algname and FB.
We report performance differences between the symmetric variant of \algname and the symmetric-contrastive variant in Fig.~\ref{fig:state_encoder} (\emph{right}). Similarly to results from Fig.~\ref{fig:wr}, we observe that gaps in performance between self-predictive and contrastive methods grow larger when learning directly from pixels. For a numerical comparison, see Tab.~\ref{tab:sptdff_sptd}.

\begin{algorithm}[t!]
    \caption{Symmetric \algname for zero-shot RL (\textcolor{our_blue}{latent-predictive} and \textcolor{our_orange}{contrastive} variants)}\label{alg:sptd_sym}
    \begin{algorithmic}
    \footnotesize
    \State \textbf{Inputs}: Dataset $\mathcal{D}$, batch size $B$, regularization coefficient $\lambda$, networks $\pi$, $T_\phi$, $\phi$
    \State Initialize target networks: $T_\phi^{-}  \leftarrow T_\phi$, $\phi^- \leftarrow \phi$
    \While{not converged}
    \State \(\triangleright\) \textcolor{gray}{Sample training batch}
    \State $\{(s_i, a_i, s_{i}')\}_{i=1}^B \sim \mathcal{D}$, $\{z_i \}_{i = 1}^B \sim \mathcal{Z}$, $\{a'_i\}_{i=1}^B \sim \{\sg{\pi(\phi^-(s_i'),z_i)}\}_{i=1}^B$
    \vspace{5pt}
    \State \vspace{0.3pt} \(\triangleright\) \textcolor{gray}{Compute \textcolor{our_blue}{latent-predictive}/\textcolor{our_orange}{contrastive} loss}
    \State \textcolor{our_blue}{$\wh{\cL} (\phi, T_\phi) =\frac{1}{2 B} \sum_{i} \left\| T_\phi(\phi(s_i), a_i, z_i) - \sg{\phi^-(s_i')}  - \gamma \sg{T_\phi^-(\phi^-(s_i'),a_i',z_i)}  \right\|^2$}
    \State \textcolor{our_orange}{$\wh{\cL} (\phi, T_\phi) =\frac{1}{2B(B-1)} \sum_{i \neq j} \big(T_\phi(\phi(s_i), a_i, z_i)^\top \phi(s'_j) - \gamma \sg{T_\phi(\phi(s'_i), a'_i, z_i)^\top \phi(s'_j)} \big)^2$}
    \State \hskip46pt \textcolor{our_orange}{$- \frac{1}{B} \sum_i T_\phi(\phi(s_i), a_i, z_i)^\top \phi(s'_i)$}
    \State \vspace{0.3pt} \(\triangleright\) \textcolor{gray}{Compute orthonormality regularization loss}
    \State $\wh{\cL}_{\rm{REG}}(\phi) = \frac{1}{2 B (B-1)} \sum_{i \neq j} (\phi(s_i)^\top \phi(s_j))^2 - \frac{1}{B} \sum_{i}\phi(s_i)^\top \phi(s_i) $
    \State \vspace{0.3pt} \(\triangleright\) \textcolor{gray}{Compute actor loss}
    \State $\{\hat{a}_i\}_{i=1}^{B} \sim \{\pi(\phi(s_i),z_i)\}_{i=1}^B$
    \State $\wh{\cL}_{\rm{actor}} (\pi) = -\frac{1}{B}\sum_{i=1}^B T_\phi(\phi(s_i), \hat a_i, z_i)^\transp z_i $
    \State \vspace{0.1pt} Update $\phi$, $T_\phi$ to minimize $\wh{\cL} (\phi, T^\phi) + \lambda \wh{\cL}_{\rm{REG}}(\phi)$
    \State Update $\pi$ to minimize $\wh{\cL}_{\rm{actor}} (\pi)$
    \State Update target networks $\phi^-$, $T_\phi^{-}$ via EMA of $\phi$, $T_\phi$
    \EndWhile
    \end{algorithmic}
\end{algorithm}

\subsection{Additional fast adaptation results}
\label{app:exp.adapt}

\begin{figure}
    \centering
    \begin{minipage}{0.9\textwidth}
    \includegraphics[width=\textwidth]{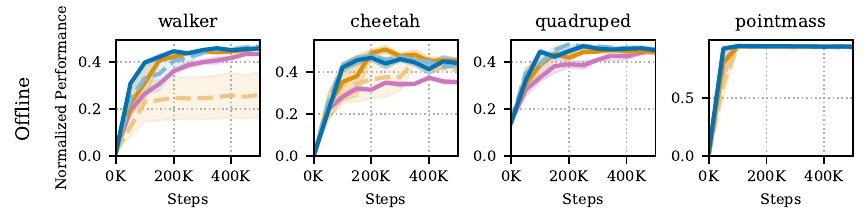}
    \end{minipage}\\
    \vspace{-6pt}
    \begin{minipage}{0.9\textwidth}
    \includegraphics[width=\textwidth]{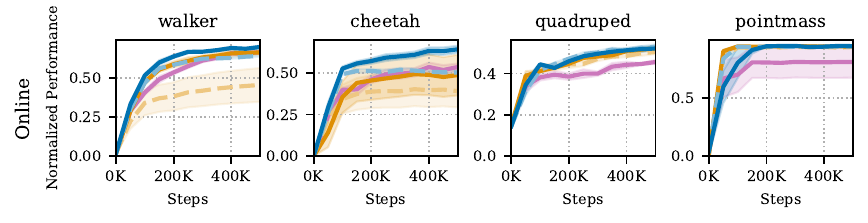}
    \end{minipage}\\
    \vspace{-4pt}
    \centering
    \rule{300pt}{1pt}\\
    \begin{minipage}{0.9\textwidth}
    \includegraphics[width=\textwidth]{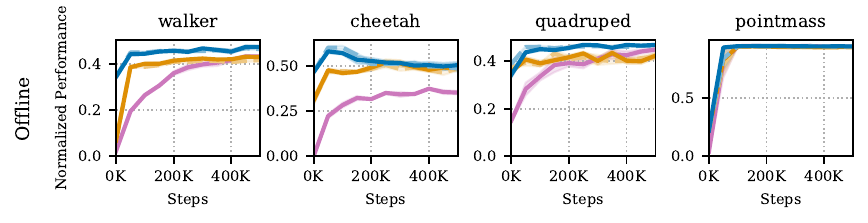}
    \end{minipage}\\
    \vspace{-6pt}
    \begin{minipage}{0.9\textwidth}
    \includegraphics[width=\textwidth]{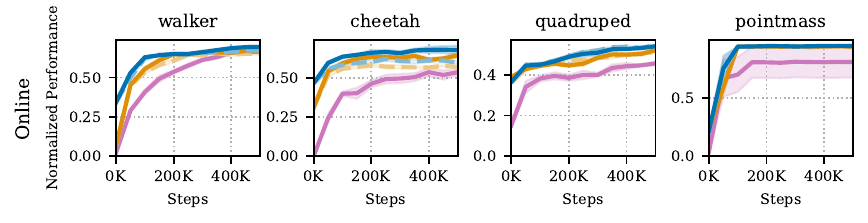}
    \end{minipage}\\
    \vspace{-4pt}
    \centering
    \rule{300pt}{1pt}\\
    \begin{minipage}{0.7\textwidth}
    \includegraphics[width=\textwidth]{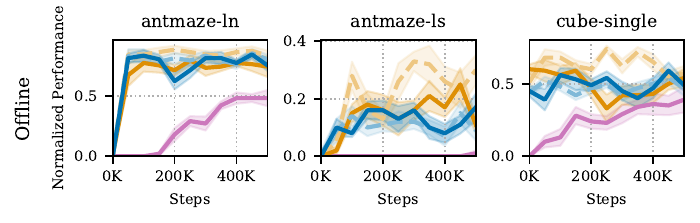}
    \end{minipage}\\
    \vspace{-6pt}
    \begin{minipage}{0.7\textwidth}
    \includegraphics[width=\textwidth]{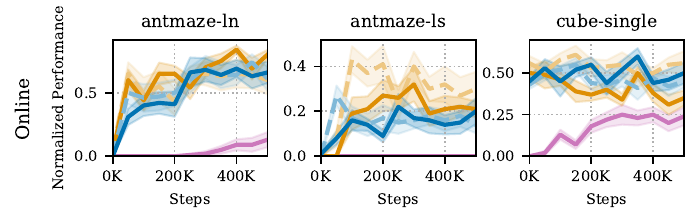}
    \end{minipage}\\
    \vspace{-4pt}
    {
    \footnotesize
    \centering
    \raisebox{0.4ex}{\textcolor{our_blue}{\rule{10pt}{2pt}}} \; \algname \quad
    \raisebox{0.4ex}{\textcolor{our_bluel}{\rule{4pt}{2pt}\,\rule{4pt}{2pt}}} \; \algname (frozen) \quad
    \raisebox{0.4ex}{\textcolor{our_orange}{\rule{10pt}{2pt}}} \; FB \quad
    \raisebox{0.4ex}{\textcolor{our_orangel}{\rule{4pt}{2pt}\,\rule{4pt}{2pt}}} \; FB (frozen) \quad
    \raisebox{0.4ex}{\textcolor{our_pink}{\rule{10pt}{2pt}}} \; Scratch \quad
    }

    \vspace{-4pt}
    \caption{Additional fast adaptation results, obtained when only loading encoder's weights (\textbf{top}), when only freezing convolutional layers (\textbf{middle}) and in OGBench (\textbf{bottom}).}
    \vspace{-4pt}
    \label{fig:fast-adapt2}
\end{figure}

We extend the empirical evaluation on fast adaptation in Section \ref{sec:exp} through Fig.~\ref{fig:fast-adapt2}. In the top part, we repeat the evaluation of Fig.~\ref{fig:fast-adapt}, while only initializing encoders to pre-trained weights. The remaining components of actor and critic need to be learned from scratch. These experiments thus recall the standard setting in which pre-trained visual representations are evaluated \citep{nair2022r3m, ma2022vip, majumdar2023we}. We observe that, while initial performance is near-zero for all methods, pre-trained representations maintain their effectiveness in terms of sample efficiency.
In the middle row, we again repeat the evaluation from Fig.~ \ref{fig:fast-adapt}, but only freeze convolutional weights for the variants represented by a dashed line. We observe that this causes the performance gap to full fine-tuning to shrink significantly, suggesting that convolutional filters extract suitable representations for both \algname and FB.
Finally, we extend our empirical evaluation to OGBench in the bottom part of Figure \ref{fig:fast-adapt2}, in which we present results for the most challenging tasks in three representative domains. This evaluation differs from the previous ones, as we found it to require strong BC regularization, even during fine-tuning.
We confirm that pre-trained representations remain beneficial in terms of sample efficient adaptation. Interestingly, we find that frozen representations pre-trained may outperform full fine-tuning in \texttt{antmaze-ls}, while they remained a bottleneck in DMC. \looseness -1

\newpage

\begin{figure}
    \centering
    \vspace{-8pt}
    \begin{minipage}{0.9\textwidth}
    \includegraphics[width=\textwidth]{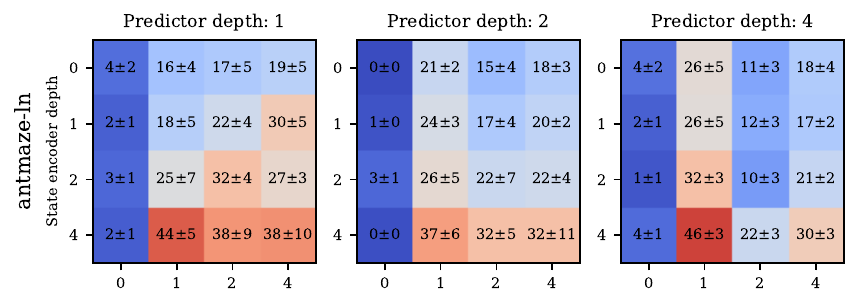}        
    \end{minipage}\\
    \vspace{-8pt}
    \begin{minipage}{0.9\textwidth}
    \includegraphics[width=\textwidth]{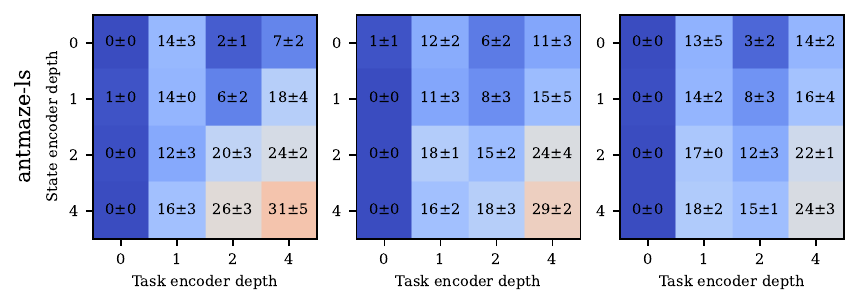}        
    \end{minipage}
    \caption{Zero-shot performance of \algname in \texttt{antmaze-ln} (top) and \texttt{antmaze-ls} (bottom) as the number of hidden layers in the encoders and predictors varies (from $0$ to $4$ and from $1$ to $4$, respectively). \looseness -1}
    \label{fig:archi}
\end{figure}

\subsection{Architectural ablations}

Architectural choices are often crucial for self-predictive learning, and capacity is usually carefully distributed between encoder and predictor \citep{guo2022byol}. This section ablates the width of the three main components in \algname in a controlled setting, namely in two visual OGBench tasks (\texttt{antmaze-ln} and \texttt{antmaze-ls}), which were chosen due to their complexity, and the fact that they often reward different approaches (see Tab.~\ref{tab:main.results}).
We measure zero-shot performance as widths change in Fig.~\ref{fig:archi}. In general, we observe that the state encoder (on the $y$ axis) should be as deep as possible (although this trend is not present in DMC, see Table \ref{tab:state_encoder}).
Instead, the task encoder (on the $x$ axis) should only be as deep as needed: at least 1 layer in \texttt{antmaze-ln} and closer to $4$ for \texttt{antmaze-ls}. Finally, we observe that the predictor may be shallow, as long as the encoders have sufficient capacity. This result matches the general conjecture that latent-predictive representations should capture the key aspects of the input, to the extent that a prediction problem, e.g. predicting successor features, may be solved with limited capacity.

\begin{figure}
    \centering
    \includegraphics[width=\textwidth]{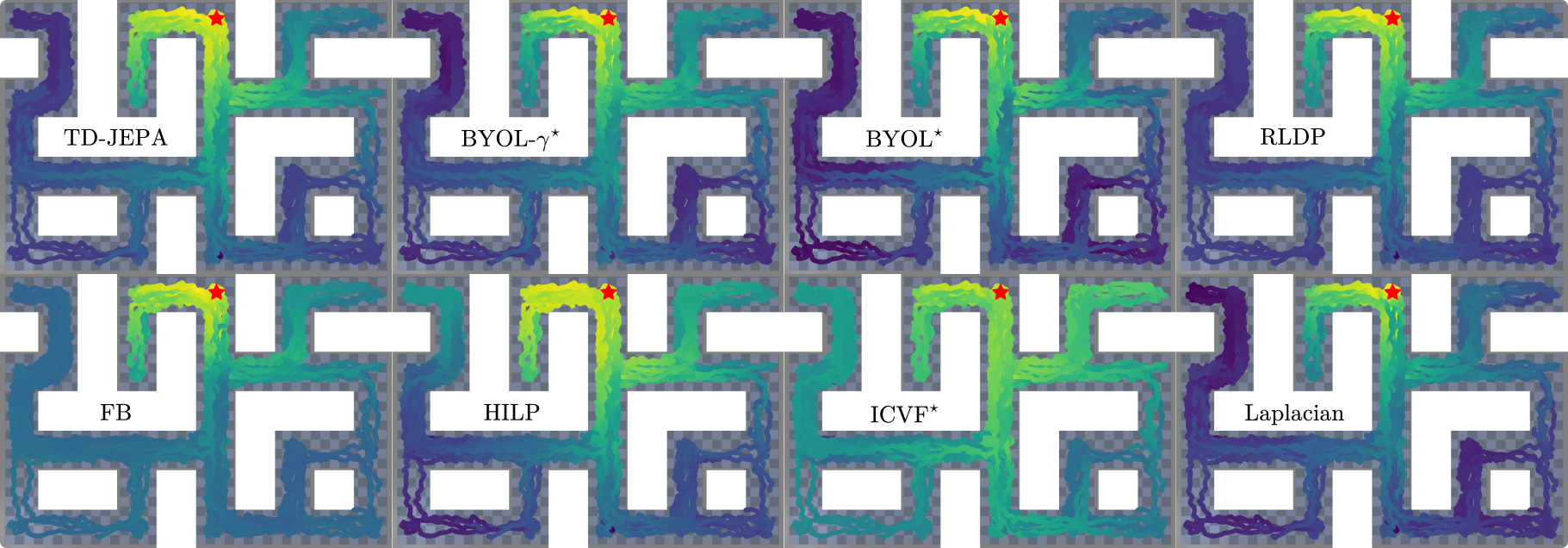}
    \vspace{-15pt}
    \caption{Cosine similarities between successor features and features (e.g., $\langle T^\phi(\phi(\cdot), a, z_g), \psi(g) \rangle$ for \algname) projected over $xy$ position of the agent's center of mass. The red star marks $g$. Similarities reflect shortest-path distances in the MDP.}
    \label{fig:similarity}
\end{figure}

\subsection{Visualization of \algname representations}

Given a state-action pair $(s,a)$ and a further state $g$, one may easily evaluate the successor measure in $g$ of the policy that tries to reach $g$ from $s$: for instance, in the case of \algname, $M^{\pi_{\phi(g)}}(g \mid s, a) \approx T^\phi(\phi(s), a, z_g)^\top \psi(g)$, where $z_g = \bE_{s \sim \mathcal{D}_{\text{rwd}}}[\psi(s)\psi(s)^\transp]^{-1} \psi(g)$. Intuitively, this is connected to how quickly the policy may reach the goal, and should reflect the dynamics of the MDP. As often done in the literature \citep{lawson2025self}, we consider the methods evaluated in Tab.~\ref{tab:main.results} and visualize these estimates for visual \texttt{antmaze-ln} in Fig.~\ref{fig:similarity}, highlighting how representations reflect temporal distances in the environment. While all plots reveal a similar pattern (which is not surprising as successor features are temporally-consistent representations by definition), some methods show inconsistent temporal distances. For instance, BYOL, RLDP, and HILP have latents of few states far from the goal with higher similarity to the latter than closer states. This is a sign of poor performance, as the predictor would be optimistic in predicting the number of steps to reach those goals. \looseness -1 

\subsection{Numerical results for performance difference plots}
\label{app:exp.numerical}

We supplement the performance difference plot in Fig.~\ref{fig:sptd-vs-spr-vs-symm} (right) and Fig. ~\ref{fig:state_encoder} (right) with Tab.~\ref{tab:sptdff_sptd}, detailing numerical results.

\begin{table}
\centering
\begin{tabular}{lccc}
\toprule
 & $\text{C-TD-JEPA}_{sym}$ & $\text{TD-JEPA}_{sym}$ & TD-JEPA \\ \midrule
DMC$_{\text{RGB}}$ (\texttt{avg}) & 437.2 \scriptsize{$\pm$ 9.8} & 598.1 \scriptsize{$\pm$ 5.9} & \textbf{628.8 \scriptsize{$\pm$ 5.5}} \\
\quad \texttt{walker} & \scriptsize{413.2} \scriptsize{$\pm$ 16.1} & \scriptsize{685.9} \scriptsize{$\pm$ 14.8} & \textbf{\scriptsize{738.9} \scriptsize{$\pm$ 3.5}} \\
\quad \texttt{cheetah} & \scriptsize{517.9} \scriptsize{$\pm$ 31.3} & \scriptsize{688.0} \scriptsize{$\pm$ 7.2} & \textbf{\scriptsize{706.0} \scriptsize{$\pm$ 4.1}} \\
\quad \texttt{quadruped} & \scriptsize{428.3} \scriptsize{$\pm$ 21.6} & \textbf{\scriptsize{606.7} \scriptsize{$\pm$ 20.1}} & \textbf{\scriptsize{626.7} \scriptsize{$\pm$ 13.6}} \\
\quad \texttt{pointmass} & \scriptsize{389.5} \scriptsize{$\pm$ 17.2} & \scriptsize{411.6} \scriptsize{$\pm$ 13.9} & \textbf{\scriptsize{443.7} \scriptsize{$\pm$ 10.9}} \\
\midrule
DMC (\texttt{avg}) & 586.3 \scriptsize{$\pm$ 14.6} & \textbf{657.5 \scriptsize{$\pm$ 3.6}} & \textbf{661.2 \scriptsize{$\pm$ 6.3}} \\
\quad \texttt{walker} & \scriptsize{757.8} \scriptsize{$\pm$ 12.4} & \textbf{\scriptsize{800.7} \scriptsize{$\pm$ 4.7}} & \scriptsize{785.2} \scriptsize{$\pm$ 6.7} \\
\quad \texttt{cheetah} & \scriptsize{583.3} \scriptsize{$\pm$ 23.3} & \scriptsize{618.1} \scriptsize{$\pm$ 11.3} & \textbf{\scriptsize{688.7} \scriptsize{$\pm$ 6.7}} \\
\quad \texttt{quadruped} & \scriptsize{565.5} \scriptsize{$\pm$ 14.4} & \textbf{\scriptsize{731.7} \scriptsize{$\pm$ 17.3}} & \scriptsize{691.4} \scriptsize{$\pm$ 5.0} \\
\quad \texttt{pointmass} & \scriptsize{438.8} \scriptsize{$\pm$ 24.5} & \textbf{\scriptsize{479.6} \scriptsize{$\pm$ 11.1}} & \textbf{\scriptsize{479.3} \scriptsize{$\pm$ 23.6}} \\
\midrule
OGBench$_{\text{RGB}}$ (\texttt{avg}) & 33.93 \scriptsize{$\pm$ 0.67} & 39.74 \scriptsize{$\pm$ 0.64} & \textbf{41.34 \scriptsize{$\pm$ 0.45}} \\
\quad \texttt{antmaze-mn} & \scriptsize{93.80} \scriptsize{$\pm$ 1.05} & \textbf{\scriptsize{95.80} \scriptsize{$\pm$ 1.09}} & \textbf{\scriptsize{96.67} \scriptsize{$\pm$ 1.11}} \\
\quad \texttt{antmaze-ln} & \scriptsize{68.00} \scriptsize{$\pm$ 3.46} & \textbf{\scriptsize{77.80} \scriptsize{$\pm$ 4.22}} & \textbf{\scriptsize{74.60} \scriptsize{$\pm$ 3.35}} \\
\quad \texttt{antmaze-ms} & \scriptsize{61.20} \scriptsize{$\pm$ 3.00} & \textbf{\scriptsize{82.20} \scriptsize{$\pm$ 1.80}} & \textbf{\scriptsize{84.40} \scriptsize{$\pm$ 3.85}} \\
\quad \texttt{antmaze-ls} & \scriptsize{8.80} \scriptsize{$\pm$ 0.90} & \scriptsize{20.60} \scriptsize{$\pm$ 1.03} & \textbf{\scriptsize{28.80} \scriptsize{$\pm$ 2.50}} \\
\quad \texttt{antmaze-me} & \textbf{\scriptsize{1.40} \scriptsize{$\pm$ 0.52}} & \textbf{\scriptsize{0.60} \scriptsize{$\pm$ 0.31}} & \scriptsize{0.20} \scriptsize{$\pm$ 0.20} \\
\quad \texttt{cube-single} & \textbf{\scriptsize{64.00} \scriptsize{$\pm$ 5.70}} & \textbf{\scriptsize{69.00} \scriptsize{$\pm$ 2.62}} & \textbf{\scriptsize{67.80} \scriptsize{$\pm$ 3.67}} \\
\quad \texttt{cube-double} & \textbf{\scriptsize{2.00} \scriptsize{$\pm$ 0.67}} & \scriptsize{1.40} \scriptsize{$\pm$ 0.60} & \textbf{\scriptsize{3.00} \scriptsize{$\pm$ 0.91}} \\
\quad \texttt{scene} & \scriptsize{4.00} \scriptsize{$\pm$ 0.79} & \scriptsize{8.00} \scriptsize{$\pm$ 1.94} & \textbf{\scriptsize{14.20} \scriptsize{$\pm$ 2.22}} \\
\quad \texttt{puzzle-3x3} & \textbf{\scriptsize{2.20} \scriptsize{$\pm$ 0.76}} & \textbf{\scriptsize{2.22} \scriptsize{$\pm$ 0.78}} & \textbf{\scriptsize{2.40} \scriptsize{$\pm$ 0.83}} \\
\midrule
OGBench (\texttt{avg}) & 35.58 \scriptsize{$\pm$ 0.97} & 35.20 \scriptsize{$\pm$ 0.49} & \textbf{37.98 \scriptsize{$\pm$ 0.77}} \\
\quad \texttt{antmaze-mn} & \textbf{\scriptsize{82.20} \scriptsize{$\pm$ 2.87}} & \scriptsize{76.40} \scriptsize{$\pm$ 2.65} & \scriptsize{70.40} \scriptsize{$\pm$ 3.72} \\
\quad \texttt{antmaze-ln} & \textbf{\scriptsize{57.20} \scriptsize{$\pm$ 4.47}} & \scriptsize{43.60} \scriptsize{$\pm$ 2.44} & \textbf{\scriptsize{57.20} \scriptsize{$\pm$ 4.25}} \\
\quad \texttt{antmaze-ms} & \textbf{\scriptsize{74.00} \scriptsize{$\pm$ 2.27}} & \scriptsize{62.80} \scriptsize{$\pm$ 3.59} & \scriptsize{61.56} \scriptsize{$\pm$ 4.53} \\
\quad \texttt{antmaze-ls} & \textbf{\scriptsize{43.20} \scriptsize{$\pm$ 3.16}} & \textbf{\scriptsize{41.40} \scriptsize{$\pm$ 3.82}} & \textbf{\scriptsize{40.60} \scriptsize{$\pm$ 2.51}} \\
\quad \texttt{antmaze-me} & \textbf{\scriptsize{21.80} \scriptsize{$\pm$ 3.30}} & \textbf{\scriptsize{17.00} \scriptsize{$\pm$ 2.65}} & \textbf{\scriptsize{20.20} \scriptsize{$\pm$ 2.39}} \\
\quad \texttt{cube-single} & \scriptsize{16.80} \scriptsize{$\pm$ 2.15} & \scriptsize{20.00} \scriptsize{$\pm$ 2.17} & \textbf{\scriptsize{34.20} \scriptsize{$\pm$ 2.88}} \\
\quad \texttt{cube-double} & \textbf{\scriptsize{3.60} \scriptsize{$\pm$ 1.07}} & \textbf{\scriptsize{2.40} \scriptsize{$\pm$ 0.78}} & \textbf{\scriptsize{3.60} \scriptsize{$\pm$ 0.78}} \\
\quad \texttt{scene} & \scriptsize{16.80} \scriptsize{$\pm$ 1.31} & \textbf{\scriptsize{39.40} \scriptsize{$\pm$ 2.17}} & \textbf{\scriptsize{38.44} \scriptsize{$\pm$ 1.37}} \\
\quad \texttt{puzzle-3x3} & \scriptsize{4.60} \scriptsize{$\pm$ 1.19} & \textbf{\scriptsize{13.80} \scriptsize{$\pm$ 1.47}} & \textbf{\scriptsize{15.60} \scriptsize{$\pm$ 1.11}} \\
\bottomrule
\end{tabular}
\caption{Performance of \algname and symmetric variants (contrastive and latent-predictive) in DMC (returns) and OGBench (success rate) with either proprioception or RGB inputs. We report means and standard errors across seeds. Numbers are bold for top algorithms if confidence intervals overlap.}
\label{tab:sptdff_sptd}
\end{table}

\section{Implementation Details}\label{app:impl}

We organize the discussion of implementation details in several subsections.

\subsection{Environments}

Zero-shot results in Section \ref{sec:exp} consider the standard Deepmind Control Suite (DMC, \citet{tassa2018deepmind}) evaluation over four domains from \citep{touati2023does}: \texttt{walker}, \texttt{cheetah}, \texttt{quadruped} and \texttt{maze}. The first three define several locomotion tasks (e.g. \texttt{walk}, \texttt{run}, \texttt{flip}), while the latter evaluates both goal- and reward-based tasks. We additionally extend the zero-shot evaluation to the more recent OGBench suite \citep{park2024foundation}, which focuses exclusively on goal reaching. For computational reasons, we consider nine representative domains. We thus evaluate navigation across five \texttt{antmaze} datasets (\texttt{medium-\{navigate, stitch, explore\}} and \texttt{large-\{navigate, stitch\}}) \footnote{For compactness, we use \texttt{mn}, \texttt{ms}, \texttt{me}, \texttt{ln}, \texttt{ln} as dataset abbreviations, respectively.}, and manipulation through \texttt{cube-\{single, double\}}, \texttt{scene} and \texttt{puzzle-3x3}. In each domain we consider the five default tasks; for consistency with DMC, we carry out zero-shot evaluation through the standard reward inference procedure, and thus define each task through its reward function. During inference, we shift each reward by $+1$, which we found to significantly improve zero-shot performance.

\subsection{Learning from pixels}

In visual experiments, environment return states as $64\times64$ RGB images. In order to alleviate non-Markovianity, states are composed by stacking $3$ frames. Each image is scaled to $[-0.5, 0.5]$ and undergoes random shifts with a maximum intensity of 2 pixels, which is slightly milder than common strategies \citep{yarats2021mastering}. The images are passed through convolutional encoders before being processed by further networks. Namely, we instantiate two convolutional encoders: one for the state encoder, and one for the task encoder (except for the symmetric variant of \algname, which only uses one). We however found that using a single, shared convolutional encoder does not significantly impact performance, and do not exclude that further specialization of the encoder might improve it.
Each convolutional encoder uses the DrQ-v2 architecture introduced in \citet{yarats2021mastering}; we briefly experimented with IMPALA- \citep{espeholt2018impala} and Dreamer-like \citep{hafner2019dream} architectures, and we found them to achieve similar results when well-tuned.

\begin{figure}
    \centering
    \vspace{-3mm}
    \includegraphics[width=\linewidth]{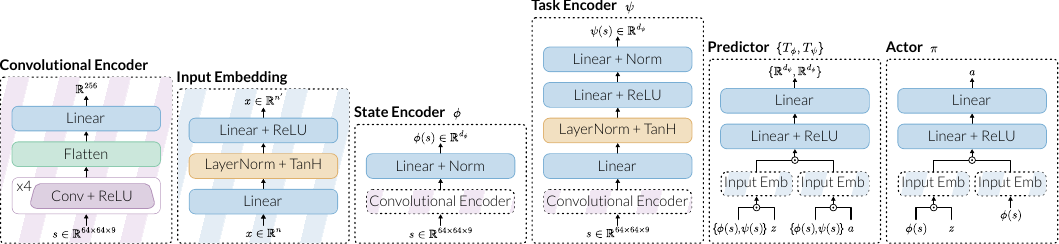}
    \vspace{-5mm}
    \caption{Overview of the architectures used by \algname in DMC$_{\rm RGB}$. We refer to Tab.~\ref{tab:hyperparameters} for the different instantiations in other domains.}
    \vspace{-5mm}
    \label{fig:archi_desc}
\end{figure}

\subsection{Architectures}

\begin{table}
\centering
\begin{tabular}{lcccc}
\toprule
  & DMC$_\text{RGB}$ & DMC & OGBench$_\text{RGB}$ & OGBench \\
 \midrule
 SFs: $T_\phi$, $T_\psi$, $F$ - hidden layers & 3 & 3 & 4 & 4 \\ 
 SFs: $T_\phi$, $T_\psi$, $F$ - hidden width & 1024 & 1024 & 512 & 512 \\ 
 State Encoder: $\phi$ - hidden layers & 0 & 0 & 0 & 4 \\ 
 State Encoder: $\phi$ - hidden width & 256 & 256 & 512 & 512 \\ 
 State Encoder: $\phi$ - output dimension $d_\phi$ & 256 & 256 & 256 & 256 \\ 
 Task Encoder: $\psi$ - hidden layers & 2 & 2 & 4 & 4 \\ 
 Task Encoder: $\psi$ - hidden witdh & 256 & 256 & 512 & 512 \\ 
 Task Encoder: $\psi$ - output dimension $d_\psi$ & 50 & 50 & 50 & 50 \\
 Actor: $\pi$ - hidden layers & 3 & 3 & 4 & 4 \\ 
 Actor: $\pi$ - hidden witdh & 256 & 256 & 512 & 512 \\ 
 \midrule
 Discount factor $\gamma$ & 0.98 & 0.98 & 0.99 & 0.99 \\
 Total gradient steps & 2M & 2M & 1M & 1M \\
 Batch size $B$ & 512 & 1024 & 256 & 256 \\
 Default learning rate & $10^{-4}$ & $10^{-4}$ & $10^{-4}$ & $10^{-4}$ \\
 Default regularization coefficient $\lambda$ & 1.0 & 1.0 & 1.0 & 1.0 \\ 
 EMA coefficient & 0.001 & 0.001 & 0.005 & 0.005 \\
 $p_{\text{goal}}$ & 0.5 & 0.5 & 0.5 & 0.5 \\
 \bottomrule
 \end{tabular}
 \caption{Architectural (top) and training (bottom) hyperparameters.}
 \label{tab:hyperparameters}
\end{table}

All algorithms rely on three types of networks, each of which is instantiated to the standard architectures from \citep{touati2023does}: (i) successor feature estimators, predictors and $F$-networks are MLPs with two layer-normalized embedding layers , (ii) state/task encoders and $B$-networks are standard MLPs with L2-normalized output and (iii) actor networks are Gaussian MLPs with a similar architecture to predictors, and fixed standard deviation of $0.2$. All networks use ReLU activations, except for embedding layers, which use TanH. The number and width of layers in DMC closely follow those described in \citet{touati2023does}, while we utilize deeper, narrower networks in OGBench, in order to better align with the implementation of benchmarked methods in \citet{Park25ogbench}.
An overview of the architectures used for \algname in DMC$_{\rm RGB}$ is depicted in Fig.~\ref{fig:archi_desc}; further architectural hyperparameters describing depth and width of these networks are reported in the first half of Table \ref{tab:hyperparameters}. We remark that the state encoder's depth in proprioception was found to be quite impactful, and was tuned according to baseline performance, as shown in Table \ref{tab:state_encoder}.

\subsection{Training hyperparameters}

The second part of Table \ref{tab:hyperparameters} describes hyperparameters used for training, and is complemented by the following discussion of further details. We again closely follow the default hyperparameters from \citet{touati2023does} and \citet{Park25ogbench} whenever possible; the batch size is reduced in visual domains to meet computational limitations. All networks are optimized through Adam \citep{kingma2014adam}. TD-targets and value estimates for SVG-like policy updates are computed as the \emph{mean} of twin networks; latents $z \in \mathcal{Z}$ are representations of random uniform states from the dataset $\psi(s)$ with probability $p_\text{goal}$, and are sampled from the hypersphere (i.e., $\mathcal{Z}$) otherwise \citep{touati2023does}.

\subsection{Method-specific details}

The baselines' implementation closely follows the public codebases, when available.

Our implementation of \textbf{FB} builds upon the one released in \citet{tirinzoni2025zeroshot}, which is in turn aligned with the code released by \citet{touati2023does}.

As described in \citet{jajoo2025regularized}, \textbf{RLDP} is implemented in the same framework, but the latent dynamics model and the task encoder are trained in parallel with the remaining components; gradients from the contrastive FB objective are not backpropagated through the task encoder. As in the original work, we found that multi-step prediction results in better performance, and we similarly adopt a prediction horizon of $H=5$. Given a batch of trajectories $\{(s_0^i, a_0^i, \dots, s_H^i)\}_{i=0}^{B-1}$, a task encoder $\psi$ and latent predictor $T_\text{RLDP}: \mathbb{R}^{d_\psi} \times \mathcal{A} \to  \mathbb{R}^{d_\psi}$, the loss for training task representations is thus:
\begin{equation}
    \wh{\cL}_{\text{RLDP}} (T_{\text{RLDP}}, \psi) = \frac{1}{2B} \sum_{i=0}^{B-1} \sum_{t=0}^{H-1} \left\| h_t^i - \psi^-(s_t^i) \right\|^2,
\end{equation}
where $h^i_0 = \psi(s_0^i)$, $h^i_t=T_\text{RLDP}(h^i_{t-1}, a^i_{t-1})$, and $\psi^-$ is a target network.
The latent predictor is instantiated with the hyperparameters presented for the SFs architecture described in Tab.~\ref{tab:hyperparameters}.

\textbf{Laplacian} \citep{mahadevan2007proto}, \textbf{HILP} \citep{park2024foundation}, \textbf{BYOL$^\star$} \citep{grill2020bootstrap}, \textbf{BYOL-$\gamma^\star$} \citep{lawson2025self} and \textbf{ICVF$^\star$} \citep{ghosh2023reinforcement} are all implemented in a successor-feature framework: the losses proposed in the original publications are optimized to retrieve a task encoder $\psi$. Considering a batch of transitions $\{(s_i, a_i, s'_i)\}_{i=0}^{B-1}$, the feature learning objectives for Laplacian, BYOL$^\star$ and BYOL-$\gamma^\star$ are, respectively:
\begin{align}
    \wh{\cL}_{\text{Laplacian}} (\psi) & = \frac{1}{2B} \sum_{i=0}^{B-1} \left\| \psi(s_i) - \psi(s_i') \right\|^2, \\
    \wh{\cL}_{\text{BYOL}} (T_{\text{BYOL}}, \psi) & = \frac{1}{2B} \sum_{i=0}^{B-1} \left\| T_{\text{BYOL}}(\psi(s_i), a) - \psi^-(s_i') \right\|^2, \\
    \wh{\cL}_{\text{BYOL}-\gamma} (T_{\text{BYOL}}, \psi) & = \frac{1}{2B} \sum_{i=0}^{B-1} \left\| T_{\text{BYOL}}(\psi(s_i), a) - \psi^-(s_i^+) \right\|^2, \\
\end{align}
where $T_{\text{BYOL}}: \mathbb{R}^{d_\psi} \times \mathcal{A} \to \mathbb{R}^{d_\psi}$ is a jointly trained latent predictor, $s_i^+ \sim M^{\pi_\beta}(\cdot|s_i, a_i)$, $\pi_\beta$ is the behavioral policy, and the minus sign ($-$) denotes target networks. HILP and ICVF instead train representations through
\begin{align}
    \wh{\cL}_{\text{HILP}} (\psi) & = \frac{1}{2B} \sum_{i=0}^{B-1} \ell_\tau^2(-\mathbf{1}_{s_i \neq g_i}-\gamma \|\psi_-(s')-
    \psi_-(g)\| + \| \psi(s) - \psi(g) \|), \\
    \wh{\cL}_{\text{ICVF}} (T_{\text{ICVF}}, \phi, \psi) & = \frac{1}{2B} \sum_{i=0}^{B-1} |\tau - \mathbf{1}_{A_i < 0}|\big(V(s_i, g_i, y_i) - \mathbf{1}_{s_i=g_i} - \gamma V_-(s'_i, g_i, y_i)\big)^2, 
\end{align}
where $\ell_\tau^2(x) = | \tau - \mathbf{1}_{x<0} |x^2$ is an expectile regression loss with expectile $\tau$, $g_i$ and $y_i$ are goals (or intents) sampled from a mixture of future and random states as described in \citet{ghosh2023reinforcement}, $V(x, y, z) = \phi(x)^\top T_{\text{ICVF}} (\psi(z)) \psi(y)$, $T_{\text{ICVF}}: \mathbb{R}^{d_\psi} \to \mathbb{R}^{d_\psi \times d_\psi}$ is a matrix predictor and $A_i = \mathbf{1}_{s_i=y_i} + \gamma V(s'_i, y_i, y_i) - V(s_i, y_i, y_i)$.
In an unified way across baselines, universal successor feature estimators $F_\psi: \mathbb{R}^{d_\phi} \times \mathcal{A} \times \mathcal{Z} \to \mathbb{R}^{d_\psi}$ (and state encoders $\phi: \mathcal{S} \to \mathbb{R}^{d_\phi}$) are then trained through standard TD-learning over features $\psi$ by optimizing:
\begin{equation}
    \wh{\cL}_{\text{SF}}(F_\psi, \phi) = \frac{1}{2B} \sum_{i=0}^{B-1} \left\| F_\psi(\phi(s_i), a_i;z_i) - \sg{\psi^-(s_i')} - \gamma \sg{F_{\psi,-}(\phi^-(s_i'), a_i';z_i)} \right\|^2,
\end{equation}
where $a'_i \sim \pi(\phi(s'_i), z_i)$.
The latent dynamics model in BYOL$^\star$ and BYOL-$\gamma^\star$, as well as the intent-conditioned predictor from ICVF$^\star$ \citep{ghosh2023reinforcement}, share the architecture of SFs, as described in Table \ref{tab:hyperparameters}, potentially dropping the subnetworks that receive states, actions or latents $z$ as necessary. The state encoder in ICVF$^\star$ receives gradients from both the ICVF loss, and the successor feature prediction, which we found to slightly improve performance compared to only propagating gradients in either direction.

We found all zero-shot algorithms to be sensitive to certain hyperparameters, such as the strength of orthonormal regularization $\lambda$. For a fair comparison, we evaluate all algorithms over a small hyperparameter grid (6 configurations in DMC, and 4 in OGBench), and report performances for the best performing configuration for each domain. For all algorithms, we sweep over two values for the learning rate of the task encoder $\psi$: $[10^{-4}, 10^{-5}]$. A second important hyperparameter is the orthonormal regularization coefficient $\lambda$, for which optimal ranges strongly differ across algorithms. In order to avoid an excessively large sweep, the ranges were thus tuned for each algorithm on a representative subset of domains, and are reported in Table \ref{tab:orthonormal}. In general, we observe that contrastive methods prefer very strong regularization, while self-predictive methods can learn with weaker regularization in certain domains. For algorithms that do not leverage orthonormal regularization (HILP, ICVF$^\star$), we instead sweep over the likelihood of sampling random goals/intents in $[0.375, 0.5]$.

\begin{table}
\centering
\resizebox{\textwidth}{!}{
\begin{tabular}{lcccccccc}
\toprule
 & Laplacian & FB & RLDP & BYOL* & BYOL-$\gamma$* & \algname \\ \midrule
DMC$_{\text{RGB}}$ & [0.01, 0.1, 1] & [0.01, 0.1, 1] & [0.01, 0.1, 1] & [0.01, 0.1, 1] & [0.01, 0.1, 1] & [0.01, 0.1, 1] \\
DMC & [0.01, 0.1, 1] & [0.01, 0.1, 1] & [0.01, 0.1, 1] & [0.01, 0.1, 1] & [0.01, 0.1, 1] & [0.01, 0.1, 1] \\
OGBench$_{\text{RGB}}$ (nav) & [0, 1] & [100, 1000] & [100, 1000] & [0.001, 0.01] & [0.001, 0.01] & [0.1, 1] \\
OGBench (nav) & [0, 1] & [100, 1000] & [100, 1000] & [0.001, 0.01] & [0.001, 0.01] & [0.1, 1] \\
OGBench$_{\text{RGB}}$ (man) & [0, 1] & [100, 1000] & [100, 1000] & [0.01, 0.1] & [0.01, 0.1] & [1, 10] \\
OGBench (man) & [0, 1] & [100, 1000] & [100, 1000] & [0.01, 0.1] & [0.01, 0.1] & [1, 10] \\
\bottomrule
\end{tabular}
}
\caption{Orthonormal regularization ranges for each algorithm. (nav) and (man) indicate navigation and manipulation domains, respectively.}
\label{tab:orthonormal}
\end{table}

\subsection{Offline correction}
\label{app:impl.bc}

The standard zero-shot evaluation pipeline \citep{touati2023does} has traditionally relied on high-coverage datasets \citep{Yarats2022exorl}. OGBench \citep{Park25ogbench} represents an interesting challenge in this sense, as most datasets are expert-like, and the incomplete support over actions induces well-known offline issues \citep{kumar2020conservative}. While these problems have been previously studied in the context of zero-shot RL \citep{jeen2024zeroshot}, we find that a novel instantiation of regularization techniques for single-task RL works well in this setting. In particular, we rely on a FlowQ-like regularization scheme \citep{park2025flow}: we train a flow model to estimate the behavioral policy, and replace the Gaussian policy with a noise-conditioned, one-step policy. This policy is regularized to samples generated by the flow model through 10 integration steps. The resulting behavior cloning loss term is normalized by the mean absolute Q-value in the batch, and scaled by a regularization coefficient $\alpha$: we use $\alpha=3$ and $\alpha=0.3$ for manipulation and navigation tasks, respectively. Both networks are instantiated with the architectural hyperparameters described for $\pi$ in Table \ref{tab:hyperparameters}.
Finally, in order to be able to track BC targets, the policy is trained directly over the state space (or on the output of convolutional encoders in visual domain), instead of acting over state representations produced by state encoders $\phi(\cdot)$.

\subsection{Evaluation: zero-shot and fast adaptation}
\label{app:impl.adapt}

While zero-shot evaluation is averaged over all tasks (4-7, depending on the domain), fast adaptation is only evaluated on the hardest task per domain, estimated through average zero-shot performance of \algname and FB.
This corresponds to \texttt{run} in \texttt{walker}, \texttt{cheetah}, \texttt{quadruped} and \texttt{square} in \texttt{maze}; we evaluate task 2 in \texttt{antmaze} domains and task 4 in \texttt{cube-single}.
Fast adaptation methods that rely on pre-trained weights are initialized from the best performing zero-shot run. Behavior cloning regularization is carried over from pre-training if present, but orthonormal regularization is no longer applied. For online adaptation, the Update-to-Data ratio is fixed to 1, and $5000$ initial transitions are collected by the policy before fine-tuning. Batches are sampled from a 50-50 mixture of the relabeled pre-training dataset, and a replay buffer containing the most recent $2\cdot10^5$ transitions. \looseness -1

The procedure to extract an actor and critic that are trainable through TD3 \citep{fujimoto2018addressing} from zero-shot agents is rather direct. First, the relabeled dataset is used to infer the optimal latent $z_r$ through linear regression. Then, the critic network may then be derived from successor feature estimates: in the case of FB \citep{touati2021learning}, $Q(s, a) \approx F(\phi(s), a, \sg{z_r})^\top \sg{z_r}$; in the case of \algname the same holds as soon as $F$ is replaced by $T_\phi$. Although adaptation schemes over $\mathcal{Z}$ are possible \citep{sikchi2025fast}, $z_r$ can be kept frozen. In order to match the output of the critic to the scale of rewards, $z_r$ is scaled such that its squared L2 norm matches the maximum reward over the dataset. Finally, the actor $\pi(s)$ can be directly extracted from zero-shot policies as soon as they are conditioned on $\sg{z_r}$.

All evaluation metrics are averaged over 10 episodes in OGBench, and 20 in DMC, and computed over 10 random seeds for OGBench and 5 for DMC.

\subsection{Pseudocode for \algname}

\begin{lstlisting}[style=mypython, caption={Python-like pseudocode for TD-JEPA (simplified).}, label={lst:tdjepa}]
def train(self):
    # sample training batch
    obs, action, next_obs = self.replay_buffer.sample()
    z = self.sample_z(obs)

    # compute targets
    next_phi = self.target_phi_encoder(next_obs)
    next_psi = self.target_psi_encoder(next_obs)
    next_action = self.actor(next_phi).sample()
    target_phi = next_psi + discount * self.target_phi_predictor(next_phi, z, next_action)
    target_psi = next_phi + discount * self.target_psi_predictor(next_psi, z, next_action)
    # compute predictions
    phi = self.phi_encoder(obs)
    psi = self.psi_encoder(obs)
    pred_phi = self.phi_predictor(phi, z, action)
    pred_psi = self.psi_predictor(psi, z, action)
    jepa_loss = mse(pred_phi - target_phi.detach()) + mse(pred_psi - target_psi.detach())
    # regularize
    phi_cov = torch.matmul(phi, phi.T)
    phi_ortho_loss = - phi_cov.diag().mean() + 0.5 * phi_cov.off_diag().pow(2).mean()
    psi_cov = torch.matmul(psi, psi.T)
    psi_ortho_loss = - psi_cov.diag().mean() + 0.5 * psi_cov.off_diag().pow(2).mean()

    # compute actor loss
    actor_action = self.actor(phi.detach(), z).sample()
    actor_pred = self.phi_predictor(phi.detach(), z, actor_action)
    actor_loss = (actor_pred * z).sum(-1).mean()

    # aggregate losses and optimize
    loss = jepa_loss + self.lambda_phi * phi_ortho_loss + self.lambda_psi * psi_ortho_loss
    loss += actor_loss
    self.optimizer.zero_grad()
    loss.backward()
    self.optimizer.step()
    # update target networks
    update_target(self.target_phi_encoder, self.phi_encoder)
    update_target(self.target_psi_encoder, self.psi_encoder)
    update_target(self.target_phi_predictor, self.phi_predictor)
    update_target(self.target_psi_predictor, self.psi_predictor)
\end{lstlisting}

Pseudocode for the default variant of \algname is shown in Listing \ref{lst:tdjepa}; the output of predictors is assumed to be averaged across twin networks.

\end{document}